\documentclass[journal]{IEEEtran}
\usepackage{multirow}
\usepackage{amsmath}
\usepackage{graphicx}
\usepackage{multirow}
\usepackage[all,pdf]{xy}
\usepackage{subfigure}
\usepackage{cite}
\usepackage{booktabs}
\usepackage{color}
\usepackage{array}
\usepackage{ulem}
\usepackage{epsfig}
\usepackage{graphicx}
\usepackage{amsmath}
\usepackage{amssymb}

\ifCLASSINFOpdf
\else
\fi

\begin{document}
	
	%
	\title{Towards Understanding the Effectiveness of Attention Mechanism}
	%
	%
	%
	
	\author{ Xiang Ye, 
		Zihang He,
		Heng Wang,
		Yong Li* 
		
		\thanks{Xiang Ye is with School of Electronic Engineering, Beijing University of Posts and Telecommunications
	Beijing, China.}
		\thanks{Yong Li is with School of Electronic Engineering, Beijing University of Posts and Telecommunications
			Beijing, China.
			e-mail: yli@bupt.edu.cn.}}

	\maketitle

	\begin{abstract}	
Attention Mechanism is a widely used method for improving the performance of convolutional neural networks (CNNs)
on computer vision tasks.
Despite its pervasiveness, we have a poor understanding of what its effectiveness stems from.
It is popularly believed that its effectiveness stems from the visual attention explanation,
advocating focusing on the important part of input data rather than ingesting the entire input.
In this paper, we find that there is only a weak consistency between the attention weights of 
features and their importance. Instead, we verify the crucial role of feature map
multiplication in attention mechanism and uncover a fundamental impact of feature map multiplication
on the learned landscapes of CNNs: with the high order non-linearity brought by 
the feature map multiplication, it played a regularization role on CNNs,
which made them learn smoother and more stable landscapes near real samples compared to vanilla CNNs.
This smoothness and stability induce a more predictive and stable behavior in-between real samples,
and make CNNs generate better. Moreover, motivated by the proposed effectiveness of feature map multiplication,
we design feature map multiplication network (FMMNet) by simply replacing the
feature map addition in ResNet with feature map multiplication.
FMMNet outperforms ResNet on various datasets,
and this indicates that feature map multiplication plays a vital role in improving the
performance even without finely designed attention mechanism in existing methods. 
	\end{abstract}
	
	
	%
	\IEEEpeerreviewmaketitle

\section{Introduction}
Convolutional Neural Networks (CNNs) have made remarkable achievements in 
various computer vision tasks,
including image classification \cite{Krizhevsky_2012,ResNet_2016,Shen2016Transform,DBLP:journals/corr/WangJQYLZWT17},
object detection \cite{SSD_2016,Faster-RCNN_2015,CornerNet_Lite}, and semantic segmentation \cite{FPN,M2Det,FCN,deeplabv3}.
Since AlexNet \cite{Krizhevsky_2012}
made the breakthrough in 2012, plenty of works \cite{BN,GoogLeNet} have been continuously investigated to further
improve their performance. 
Attention Mechanism (AM) is
firstly proposed for Nature Language Processing (NLP) \cite{vaswani2017attention}, 
and later becomes one of the most effective methods for computer vision tasks \cite{DBLP:journals/corr/WangJQYLZWT17,Oktay2018AttentionUL}. 
At a high level, AM is a technique that aims to improve the performance of CNNs by
designing attention modules assigning different attention weights to input features.

The practical success of AM is indisputable and there are a number of works that
propose attention modules improving the performance for computer vision tasks \cite{DBLP:journals/corr/WangJQYLZWT17,Oktay2018AttentionUL,SE-Net,CBAM}.
Somewhat shockingly,
however, despite its prominence, 
we still have a poor understanding of what the effectiveness of
AM stems from and few the existing works seem to bring us any closer to understanding
this issue.

Currently, the most widely accepted explanation of AM's success, as well as its original motivation,
intuitively argues AM as the visual attention that advocates focusing on the
important part of input data rather than ingesting the entire input.
Informally, important features are assigned large valued
attention weights generating large valued features.
Even though this explanation is widely accepted, 
it still lacks concrete or direct evidence.
In particular, the relationship between the value of features and their importance is unclear.
The chief goal of this paper is trying to address all these shortcomings. Our exploration leads to somewhat
interesting discoveries.

\textbf{Contributions:}

1) \textbf{This paper examines the relationship between the attention weights of features and their
	importance scores evaluated by existing feature importance evaluation methods.
Our experiments however, only find a weak correlation between them, which 
makes the visual attention explanation of AM more suspectable} 

Feature importance evaluation \cite{hooker2018evaluating,ahern2019normlime,wojtas2020feature,OnInterpretability}
is a fundamental problem in deep model interpretation and
feature visualization.
Saliency methods \cite{2019FullGradientRF,2017arXiv170603825S}, one of the most popular feature importance evaluation methods,
assign to each input feature an importance score, which measures the usefulness
of that feature for the performance \cite{2019FullGradientRF}.
A large importance score of a feature means a large performance drop when the feature is unavailable.

Since AM is claimed to advocate focusing on the important part of input data,
to verify the relationship between the attention weights of features and their importance scores,
this paper checked the Kendall rank correlation scores \cite{brandenburg2013comparing} and
the cosine similarity between them. However, only a weak correlation is found,
which indicates that AM does not precisely advocate focusing on 
the important part of input data, 
and makes the visual attention explanation of AM more suspectable.

2) \textbf{CNNs equipped with AM are revisited from a functional viewpoint.
	Firstly, we demonstrate the
importance of feature map multiplication operator in AM, 
and propose that it turns
CNN from a piecewise linear function into a piecewise function of high degree,
and brings about the smooth and stable learned landscapes near real samples.
\uline{Secondly, the smooth and stable learned landscapes are demonstrated to be relevant to the generalization ability of CNNs and
are the reason for the effectiveness of AM.
Finally, we conjecture that AM plays a regularization role on the landscapes learned by CNNs which makes them
smooth and stable, and
verify it by using it in conjunction with data augmentation methods.}}

\begin{figure}[h]
	\centering 
	\includegraphics[width=0.24\textwidth]{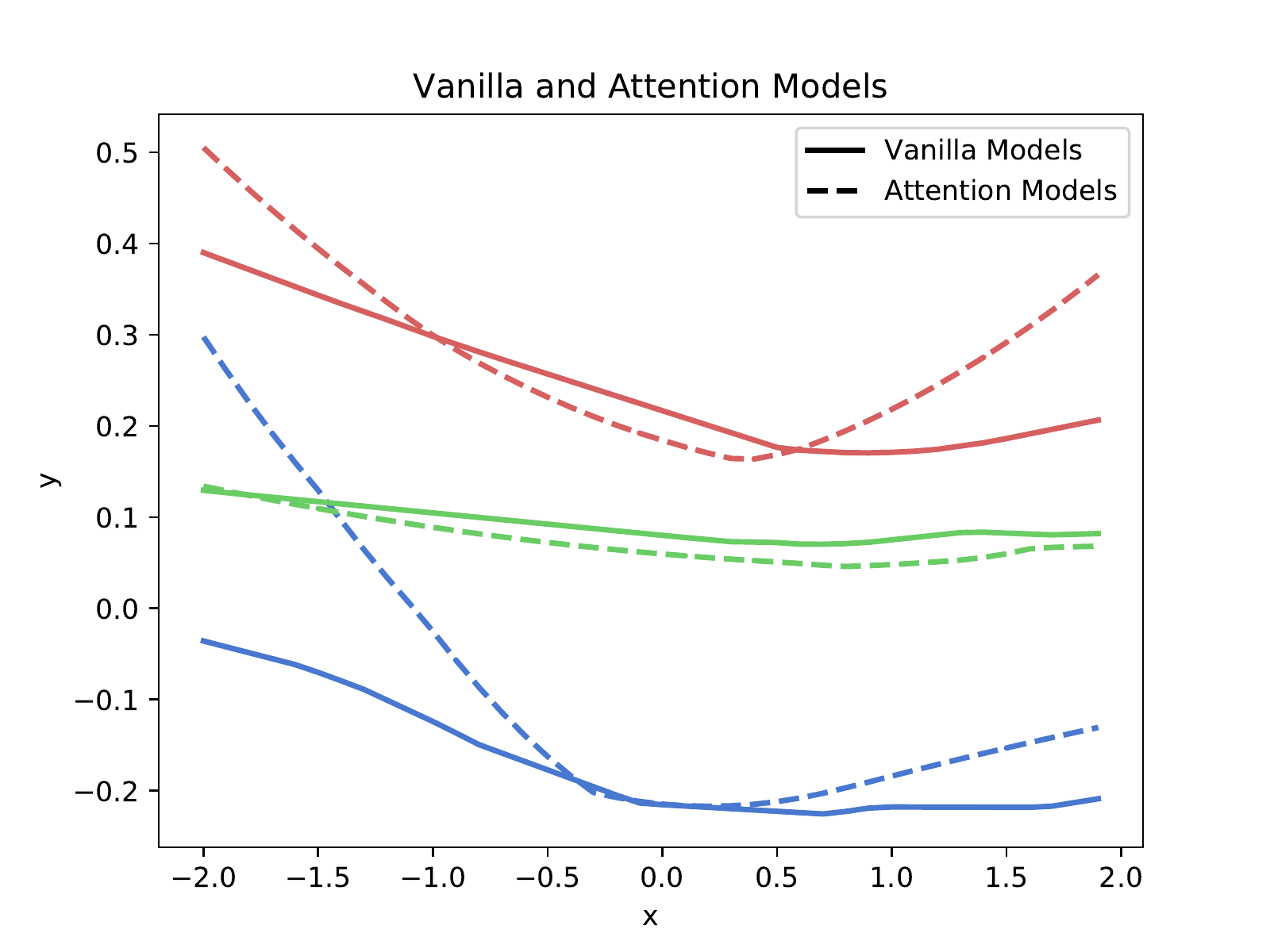}
	\includegraphics[width=0.24\textwidth]{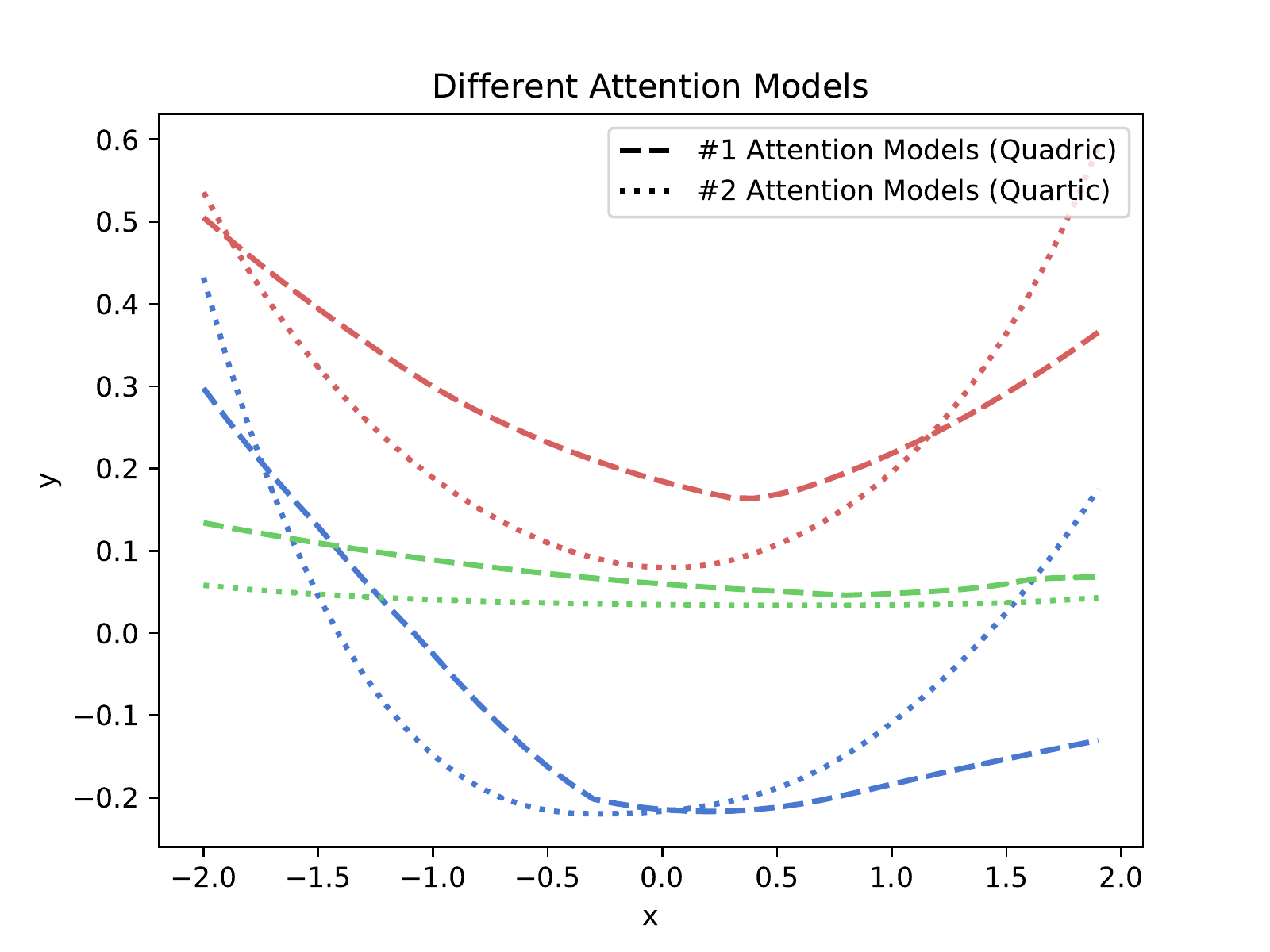}
	\caption{A simple view of the curves
		represented by vanilla neural networks (NNs)
		and attention mechanism equipped NNs
		with one-dimensional input and output. All the parameters of the depicted NNs are random filled.
		It illustrates that attention mechanism affects the property of curves represented by NNs,
		i.e., the curve represented by attention mechanism equipped NNs seems to be smoother than
		vanilla NNs.}
	\label{fig:simple} 
\end{figure}

If it is suspectable to explain AM by the importance of feature, what is a more reasonable explanation?
Few the existing works seem to bring us any closer to understanding
this issue. This paper tried to address this issue from a functional viewpoint.

Figure \ref{fig:simple} gives a simple view of the curves
represented by vanilla neural networks (NNs)
and AM equipped NNs
with one-dimensional input and output. All the parameters of the depicted NNs are random filled.
It illustrates that AM affects the property of curves represented by NNs,
i.e., the curve represented by AM equipped NNs seems to be smoother than
vanilla NNs. Formally, AM containing feature map multiplication operator
turns NNs from piecewise linear functions
into piecewise functions of high degree.

Motivated by this, we conjecture that the feature map multiplication changing the degree of
functions represented by CNNs is the key for the effectiveness of AM.
This paper verifies the conjecture empirically and explores the relationship between feature map
multiplication/AM and the generalization ability of CNNs. 
Through visualizing the curve represented by CNNs along elaborately picked one-dimensional trajectories in input space,
we find it bring about smoother and more stable learned landscapes compared to the vanilla CNNs,
which makes the AM equipped CNNs less sensitive to perturbations and generate better.

Moreover, we conjecture that feature map multiplication/AM plays a regularization role on the 
learned landscapes of CNNs. It is verified by using it in conjunction with a augmentation method, mixup \cite{zhang2018mixup}.

3) \textbf{Based on the contribution 2, to further verify the benefit of feature map multiplication operator in CNNs.
We propose a module simply changing the feature map addition operator in ResNet \cite{ResNet_2016}
into feature map multiplication operator,
named FMMNets (Feature Map Multiplication Networks). The performance of ResNet and FMMNets is compared on various datasets.}

Contribution 1 claimed that attention weights of features do not express their importance.
Contribution 2 claimed that the feature map multiplication operator bringing
about smooth and stable learned landscapes is the key for the effectiveness of AM.
Based on the two contributions, FMMNet is proposed which is the first module motivated by the 
effectiveness of feature map multiplication, not the visual attention like the existing
attention modules. The effectiveness of FMMNet will be verified.

\textit{\uline{Indeed, feature map multiplication is not only be used in AM. 
Although not motivated by the visual attention, some works have already benefit from it, e.g.,
non-linear kernels} \cite{wang2019kervolutional,zoumpourlis2017non}.}
	\textit{\uline{Despite the different motivations, for the first time, this paper regards them and
works about AM as the same kind of works, which
explores how to use feature map multiplication efficiently in CNNs.}}
A review of these works will be illustrated in Section
\ref{sec:related}.

\section{Related Works}
\label{sec:related}
Despite the different motivation, this paper for the first time
regarded all the works benefiting from feature map multiplication as the
same kind of works.

\subsection{Attention Mechanism}
Attention mechanism was an effective method for improving the performance of CNNs.
Jaderberg \textit{et al.} proposed STN (Spatial Transformer Networks)
\cite{Jaderberg_stn_2016} in 2016,
using attention mechanism in spatial dimensions to allow spatial manipulation of
data within the network. Experiment result showed that
STNs were invariant to the translations
of input images, such as scale, rotation and more generic warping, resulting state-of-the-art performance
on several benchmarks. Zhao \textit{et al.} proposed PSANet (Point-wise Spatial Attention Network) \cite{zhao2018psanet}
in 2018, which used learned spatial attention mask to connect every pixel in the feature map with 
all the other pixels in the spatial dimension. It relaxed the local neighborhood constraint of convolutional
kernels and achieved top performance on various competitive scene parsing datasets \cite{zhao2018psanet}.

Despite the spatial attention mechanism mentioned above, SE-Net (Squeeze-and-Excitation Networks) \cite{SE-Net} and
ECA-Net (Efficient Channel Attention Network) \cite{2019arXiv191003151W}
explored the attention mechanism in the channel dimension. Hu \textit{et al.} proposed SE-Net in 2017, which
used attention mechanism to build the relationship between feature map channels. SE-Net showed its effectiveness
on various computer vision tasks, such as image classification, object detection, and semantic segmentation and 
won the first place in the ILSVRC 2017 competition. Based on SE-Net, Wang \textit{et al.} proposed ECA-Net in 2020.
It was a light-weight channel attention method and experiment result showed that ECA-Net was efficient while
performing favorably against its counterparts.

Other works about attention mechanism including 3D attention modules \cite{CBAM,CSCSE}
and dynamic convolution kernels \cite{chen2020dynamic,NEURIPS2019_f2201f51,Zhang2020DyNetDC},
attention mechanism on convolution kernels.

\subsection{Non-linear Kernel}

Motivated by the non-linear operations in the response of complex visual cells,
Zoumpourlis \textit{et al.} proposed NCF \cite{zoumpourlis2017non}
in 2017. It turned linear convolution layers
into quadratic forms and thus turned CNNs from linear piecewise functions into
piecewise functions of high degree. Experiment result demonstrated that CNNs
combining linear and non-linear filters in their convolutional layers outperformed
CNNs using standard linear filters with the same architecture.
Chen \textit{et al.} proposed KNN \cite{wang2019kervolutional} in 2020,
approximating complex behaviors of human perception systems.
It generalized convolution, enhanced the model capacity, and captured higher order
interactions of features, via patch-wise kernel functions \cite{wang2019kervolutional}.
Experiment result demonstrated that KNNs achieved higher accuracy and faster convergence than
conventional CNNs.

These methods incorporated feature map multiplication operator in the convolutional layers
and turned them from linear operators into non-linear operators. This paper believes they
were essentially the same as attention mechanism, both benefit from the non-linear ability of
feature map multiplication. How to use feature map multiplication efficiently was the key for the
kind of works, i.e., by the feature map multiplication within patches like KNN \cite{wang2019kervolutional},
or on a channel/spatial dimension like attention mechanism.

\section{Does Attention Mechanism Express the Importance of Features?}
\label{sec:Importance}
\subsection{Overview}
Feature importance evaluation was beneficial to CNN interpretation
or its performance improvement.
AM (Attention Mechanism), one of the most prominent methods explained to
benefit from feature importance evaluation, was claimed to 
advocate focusing on the important part of input data.
However, while almost all the existing papers about AM admitted the explanation of AM and
were even motivated by it, it had never been verified and seemed to have little concrete evidence
supporting it. 

This paper addressed the issue and verified to what extent the attention weights of features
could represent their importance. 
To achieve this goal, this paper examined the consistency between attention weights of features
and their importance scores evaluated by feature importance evaluation methods,
e.g., saliency methods \cite{OnInterpretability}, 
identifying which attributes were most relevant to a prediction and evaluating the
importance of features.

In detail, a model equipped with SE module (Squeeze-and-Excitation module) \cite{SE-Net},
a prominent AM method,
was firstly fully trained and 
for every test sample, their attention weights were regarded as the importance scores of the
corresponding features evaluate by AM. 
Secondly, the importance scores of the corresponding features were also evaluated by
saliency methods, FullGrad \cite{2019FullGradientRF}
and SmoothFullGrad \cite{2017arXiv170603825S}.
Finally, the similarities between the attention weights of features and the
importance scores
evaluated by the saliency methods 
were compared using the Kendall rank correlation scores \cite{brandenburg2013comparing} and the cosine similarity scores.
Kendall rank correlation scores described the rank similarity between the feature importance scores
evaluated by AM and saliency methods.
The similarity scores described their cosine similarity.
They both represented the extent that AM could express the importance of a feature.

\subsection{Results}

\begin{figure}[h]
	\centering 
	\subfigure{\includegraphics[width=0.4\textwidth]{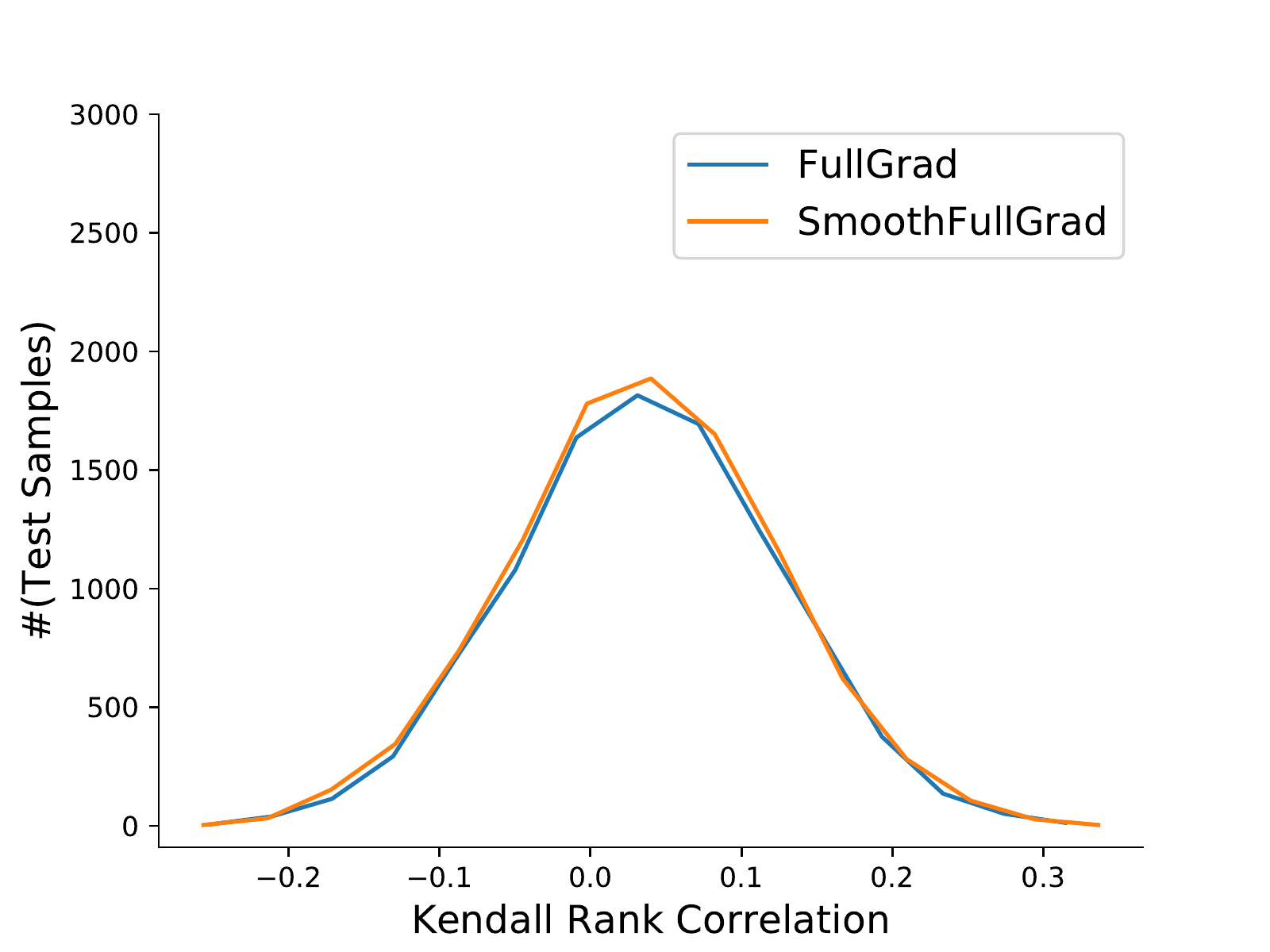}\label{fig:0.89cosine}}
	\subfigure{\includegraphics[width=0.4\textwidth]{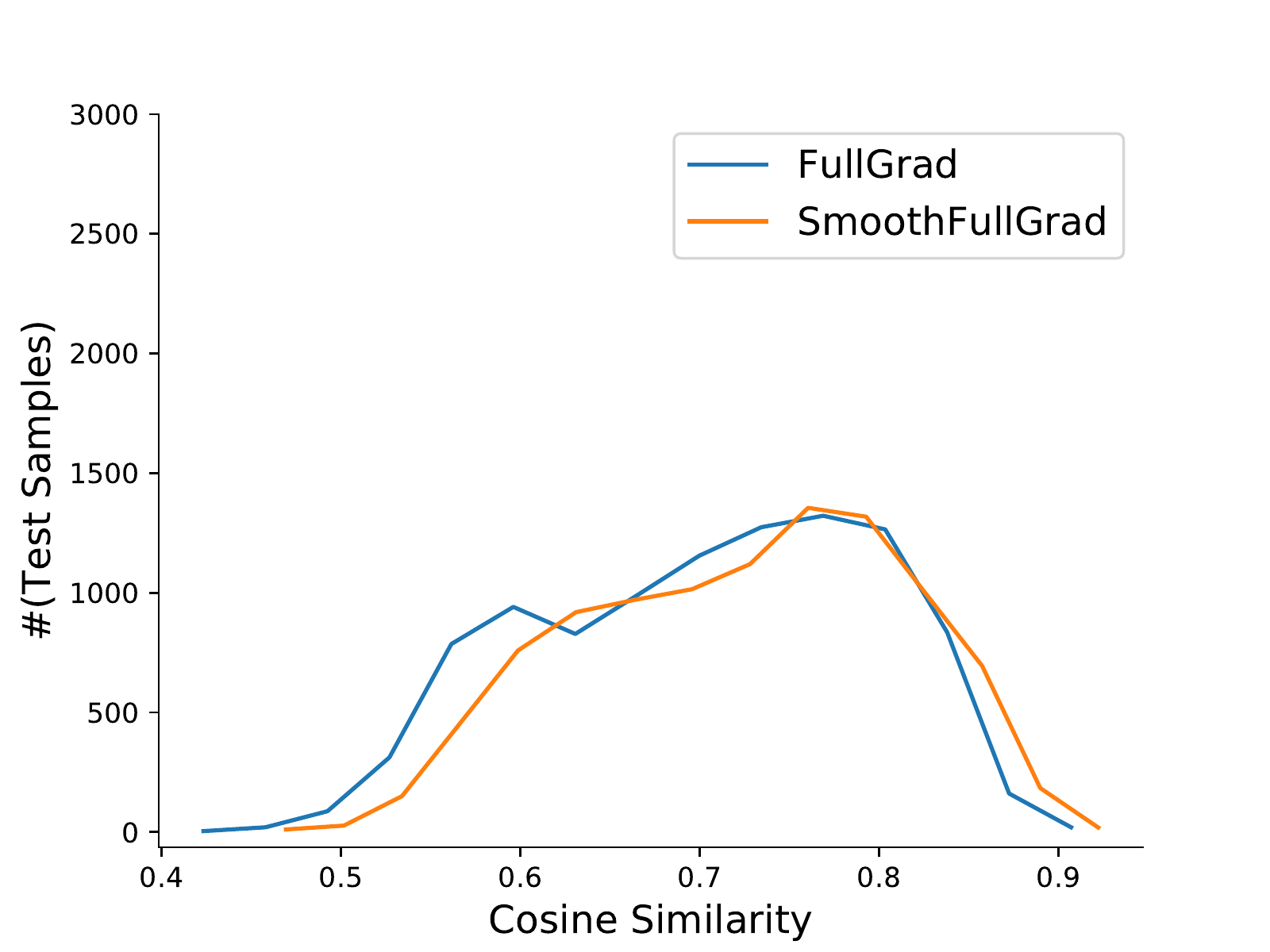}\label{fig:cosines}}
	\caption{The Kendall rank correlation scores \cite{brandenburg2013comparing} and
		the cosine similarity scores between attention weights of features and their
		feature importance 
		evaluated by saliency methods, FullGrad \cite{2019FullGradientRF}
		and SmoothFullGrad \cite{2017arXiv170603825S}, for 10k test images.
	The most test samples only yielded
	0.05 the Kendall rank correlation scores and 0.7+ cosine similarity scores.
	It indicated that in the most cases, the attention weights of features and their
	importance only exhibited a weak consistency. AM does not precisely expressed the 
	importance of features, thus does not precisely
	advocate focusing on the important part of input data. This made
	the visual attention explanation of AM more suspectable.}
	\label{fig:AwI} 
\end{figure}

Figure \ref{fig:AwI} illustrated 
the Kendall rank correlation scores \cite{brandenburg2013comparing} and
the cosine similarity scores between attention weights of features and their
feature importance 
evaluated by saliency methods, FullGrad \cite{2019FullGradientRF}
and SmoothFullGrad \cite{2017arXiv170603825S}, for 10k test images.
It denoted the extent that AM expressed the importance of
features.
For both the Kendall rank correlation \cite{brandenburg2013comparing} and
the cosine similarity, a larger score represented the 
attention weights of features were more consistent with their
feature importance.

As shown in Figure \ref{fig:AwI}, the most test samples only yielded
0.05 the Kendall rank correlation scores and 0.7+ cosine similarity scores.
It indicated that in the most cases, the attention weights of features and their
importance only exhibited a weak consistency. AM does not precisely expressed the 
importance of features, thus does not precisely
advocate focusing on the important part of input data. This made
the visual attention explanation of AM more suspectable.
If AM did not benefit from the importance of features, what its effectiveness stems from?
Section \ref{sec:enlightenment} explored this question.

\section{Enlightenment for Understanding Attention Mechanism?}
\label{sec:enlightenment}
The effectiveness of AM were explored from three aspects.
Firstly, the importance of feature map multiplication in AM was verified.
It gave the heuristic that AM benefit from the multiplication operator between feature maps in AM
turning CNN into a piecewise function of high degree.

Secondly, to explore how feature map multiplication/AM improved the performance of CNNs, 
the functional curves represented by the learned neurons along carefully chosen trajectories in input space
were visualized and compared for both AM equipped and vanilla CNNs.
The visualization results showed that AM equipped CNNs yielded smoother and more stable
landscapes near real samples compared to vanilla CNNs. 
\uline{The smooth and stable learned landscapes of AM equipped CNNs indicated they 
might be less sensitivity to the perturbations of input samples.}
It was verified by comparing the sensitivity coefficients \cite{9412496,novak2018sensitivity}
between AM equipped and vanilla CNNs, and they yielded smaller sensitivity
scores compared to vanilla CNNs.
As various
works \cite{9412496,novak2018sensitivity} demonstrated, CNNs with smaller sensitivity coefficient could generate better,
which was exactly what the smooth and stable learned landscapes of AM equipped CNNs brought about.


\uline{In the second step, the smooth and stable learned landscapes were demonstrated to be
the reason for better generation ability of AM equipped CNNs.
We conjectured that feature map multiplication/AM played a regularization role
on the learned landscapes.}
Finally, to examine the conjecture,
a data augmentation method directly manipulating
the landscapes of CNNs by minimizing the loss of augmented samples
was used. 
The results showed that with the data augmentation method,
vanilla CNNs achieved comparable performances to AM CNNs without it, which demonstrated
that it had the same benefit of the augmentation method and AM. When using the augmentation
method, AM could no longer improve the performance over vanilla CNN, which demonstrated that AM and the
augmentation method might improve the performance of vanilla CNN in the same manner.
AM played
a regularization role on the landscapes of the trained without external supervisors, which
made them smooth and stable near real samples.

\subsection{Feature Map Multiplication is the Key}
\label{sec:key}
\subsubsection{Overview}
\label{sec:key:overview}

Regarding CNNs as functions to fit the relationship between real data,
two factors could be the reasons for AM
improving the performance, the ReLU operator and the feature map multiplication operator
in the attention block.
The ReLU operators in AM increased the non-linearity of CNNs, which could be the reason that
AM equipped CNNs could fit more complex functions and performed better than vanilla CNNs.
Another reason for AM improving the performance 
could be the feature map multiplication operators which made CNNs piecewise functions of high degree
and brought about high order non-linearity to CNNs,
as assumed in Figure \ref{fig:simple}.
This paper dug the real factor for AM improving the performance and conducted two kinds of experiments 
according to the two factors.

\begin{itemize}
\item To verify the importance of ReLU, the performance of AM equipped CNNs
with or without ReLU operator in the attention block was compared.

\item To verify the importance of feature map multiplication, we compared the performance
between AM equipped CNNs with feature map addition replacing feature map
multiplication or not.
\end{itemize}

The two kinds of experiments found out the real factor in AM for improving the performance of CNNs.
It was helpful for understanding the true nature of AM.

\subsubsection{Implementation}
To find the key factor in AM, four kinds of models were trained and tested, CNNs (vanilla CNNs),
AM CNN (attention mechanism equipped CNNs), AM$_{no RL}$ CNN (attention mechanism equipped CNNs with no
ReLU operators in the attention block), and AM$_{no MTP}$ CNN (attention mechanism equipped CNNs with
the feature map addition operator replacing the feature map multiplication operator in AM).
For vanilla CNNs, we built toy CNNs with three convolutional layers and a fully-connected layer.
The number of kernels in each convolutional layer was 32, 64, and 64 respectively.
SE modules were embedded into the vanilla CNNs for 
AM CNNs, AM$_{no RL}$ CNNs, and AM$_{no MTP}$ CNNs.
All the models were trained on MNIST and CIFAR-10 datasets 
using momentum optimizer for 200 epochs with a batch size of 400
and a momentum of 0.9.

\subsubsection{Result}
\begin{table}[ht]
	\centering
	\caption{Experiment for finding the real factor in AM improving the performance.
		AM$_{no RL}$ CNN yielded a comparable performance to the AM CNN.
		It indicated that the ReLU operator in the attention block contributed little to the
		effectiveness of AM.
		However, without feature map multiplication, AM$_{no MTP}$ CNN achieved an accuracy comparable to vanilla CNN,
		which indicated that the feature map multiplication was crucial for the effectiveness of AM.
	}
	\renewcommand\arraystretch{1.2}
	\label{tab:importance}
	\begin{tabular}{c|cccc}
		\hline 
		Datasets &CNN&AM CNN&AM$_{no RL}$ CNN&AM$_{no MTP}$ CNN\\
		\hline
		MNIST&98.61&99.03&99.06&98.69\\
		CIFAR-10&79.57&80.40&80.12&79.66\\
		\hline	
	\end{tabular}
\end{table}

Table \ref{tab:importance} showed the results of finding the key factor in AM for improving
the performance of CNNs.
AM CNN outperformed vanilla CNN
on both MNIST and CIFAR-10 datasets, which verified the effectiveness of
AM.
When removing the ReLU operator in the attention block,
AM$_{no RL}$ CNN yielded a comparable performance to the AM CNN, and both improved the
performance over vanilla CNN.
It indicated that the ReLU operator in the attention block contributed little to the
effectiveness of AM and AM$_{no RL}$ CNN could improve the performance of
vanilla CNN without the ReLU activation operator in the attention block.
Thus, ReLU operator in the attention block was not the reason for the effectiveness of AM.

However, when replacing the feature map multiplication operator with the 
feature map addition operator, AM$_{no MTP}$ CNN demonstrated an accuracy drop compared with
AM CNN. It indicated that feature map multiplication was a factor for AM to improve the performance.
Without feature map multiplication, AM$_{no MTP}$ CNN achieved an accuracy comparable to vanilla CNN,
which indicated that the feature map multiplication was crucial for the effectiveness of AM.

\subsubsection{Conclusion}
Based on the result of Section \ref{sec:Importance} and \ref{sec:key},
we found that AM did not precisely express the importance of features, but benefited from
the feature map multiplication operator bringing about high order non-linearity to CNNs.
Indeed, besides AM, papers about non-linear filter
\cite{zoumpourlis2017non,wang2019kervolutional} also introduced feature map multiplication into CNNs.
It was speculated that they were the same kind of method both
benefiting from the feature map multiplication, 
despite the visual attention explanation for AM. How to use the feature map multiplication efficiently in CNNs
was the key problem for the kind of methods.

To verify the effectiveness of feature
map multiplication and to some extent refute the elaborately designed AM module based on the
visual attention motivation, we simply replaced the feature map addition operator in ResNet \cite{ResNet_2016}
with feature map multiplication operator
and proposed FMMNet. The experimental result would be illustrated
in Section \ref{sec:FMMNet}. The following section would further explore how feature map multiplication
improved the performance of CNNs.

\subsection{The Landscape Learned by Attention Mechanism Equipped CNNs and Vanilla CNNs}
\label{sec:landscape}

\subsubsection{Overview}

As illustrated above, AM was essentially the feature map multiplication in CNNs.
It brought about high order non-linearity to CNNs and
changed CNNs from piecewise linear functions to piecewise functions of high degree.
How the high order non-linearity improved the performance was still a question.

Looking into the question, 
we visualized the curve represented by the final neurons of trained CNNs
along elaborately picked one-dimensional trajectories in input space. 
The result showed that compared with the curved represented by vanilla CNNs, the
curved represented by AM CNNs were smoother and more stable, which might indicate they were
less sensitive to the perturbations of real samples. 
To verify it, two kinds of
sensitivity metrics, $\mathbf{J}$ \cite{novak2018sensitivity} and $\mathbf{S}$ \cite{9412496} 
were further compared between AM CNNs and vanilla CNNs.
It showed that with smoother
and more stable landscapes learned by AM CNNs, they yielded smaller sensitivity scores and
were less sensitivity to the perturbations of real samples compared to vanilla CNNs.
Several works \cite{9412496,novak2018sensitivity} have revealed the fact that the sensitivity of CNNs was very
related to their generalization ability, i.e., CNNs with smaller sensitivity scores could generate
better and yield better performance.
Thus, we claimed that the high order non-linearity operator, feature map multiplication in AM brought about smooth and stable
learned landscapes, which contributed to their
less sensitivity to the perturbations of input samples and made they generate better than
vanilla CNNs. 


\subsubsection{Visualization the Landscapes Learned by Attention Mechanism Equipped CNNs and Vanilla CNNs}
\label{sec:visualization}

\begin{figure}
	\centering
	\includegraphics[width=0.25\textwidth]{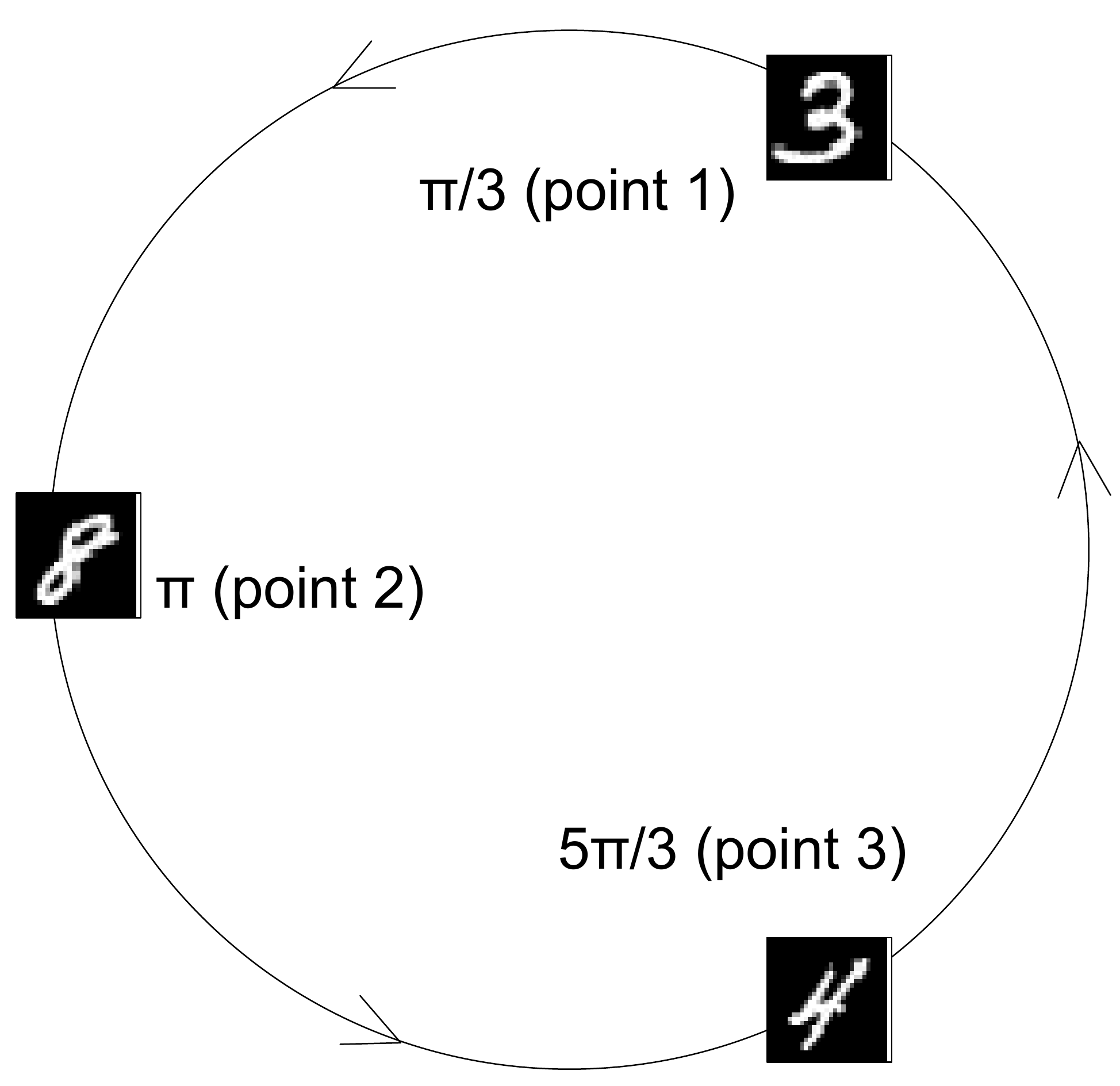}
	\caption{Cross-class ellipses. The ellipses passed through three test samples belonging to different classes
		\cite{novak2018sensitivity}. 
		The ellipses traversed samples that were linear combinations of samples belonging to different classes.
		Visualizing the curves of neurons along the ellipses demonstrated the behaviors of neurons between inputting samples
		belonging to different classes. It exhibited the boundaries of CNNs keeping its prediction of one class
		and their sensitive
		to the perturbations containing the information of samples belonging to different classes.}
		\label{fig:tr} 
\end{figure}

Feature map multiplication in AM brought about high order non-linearity to CNNs.
To explore its effect on the learned landscape, we visualized the curves of the neurons in the last layer along one-dimensional
trajectories in input space. 
Each trajectory passed through three samples picked
from the test dataset \cite{novak2018sensitivity}, to ensure it was not far away
from real samples and lay in the regions that we were interested in. Specifically, two kinds of trajectories were picked:

\begin{itemize}
	\item \textbf{Cross-class ellipses}. The ellipses passed through three test samples belonging to different classes,
	as shown in Figure \ref{fig:tr} \cite{novak2018sensitivity}. 
	The ellipses traversed samples that were linear combinations of samples belonging to different classes.
	Visualizing the curves represented by neurons along the ellipses
	demonstrated the behaviors of neurons between inputting samples
	belonging to different classes. It exhibited the boundaries of CNNs keeping its prediction of one class
	and their sensitive
	to the perturbations containing the information of samples belonging to different classes.
	For the kind of ellipses, we visualized the three neurons representing the three picked classes in the
	last layer.
		
	\item \textbf{Single-class ellipses}. The ellipses were similar to the previous one shown in Figure \ref{fig:tr}, but passed through
	three test samples belonging to the same class. Visualizing the curves of neurons along the ellipses demonstrated the behaviors
	of neurons between inputting samples belonging to the same class. It illustrated the sensitive of CNNs to the perturbations
	containing the information of samples belonging to the same classes.
	For the kind of ellipses, we visualized the neuron representing the picked class in the
	last layer.
\end{itemize}

\begin{figure*}[h]
	\centering 
	\includegraphics[width=0.24\textwidth]{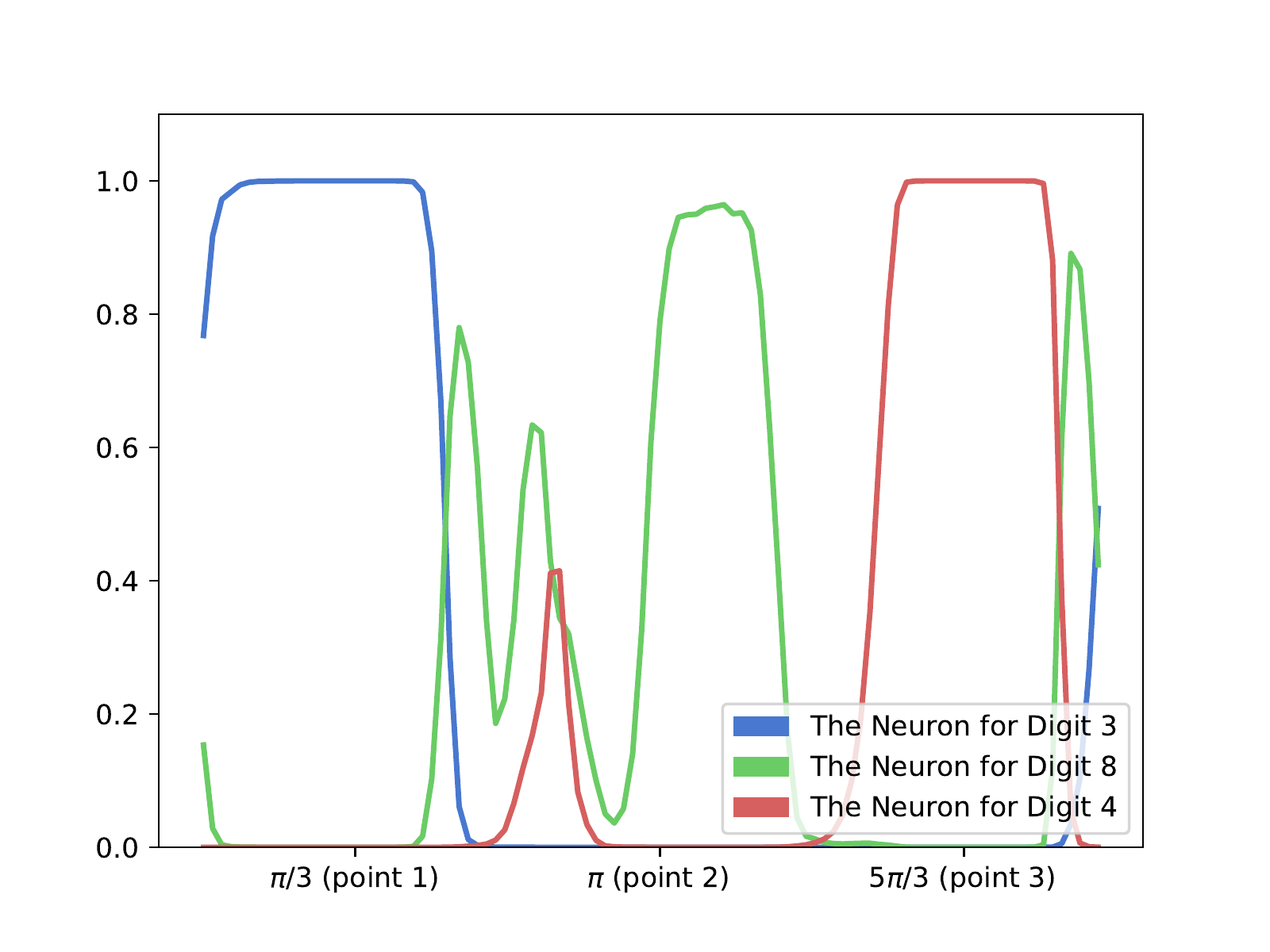}
	\includegraphics[width=0.24\textwidth]{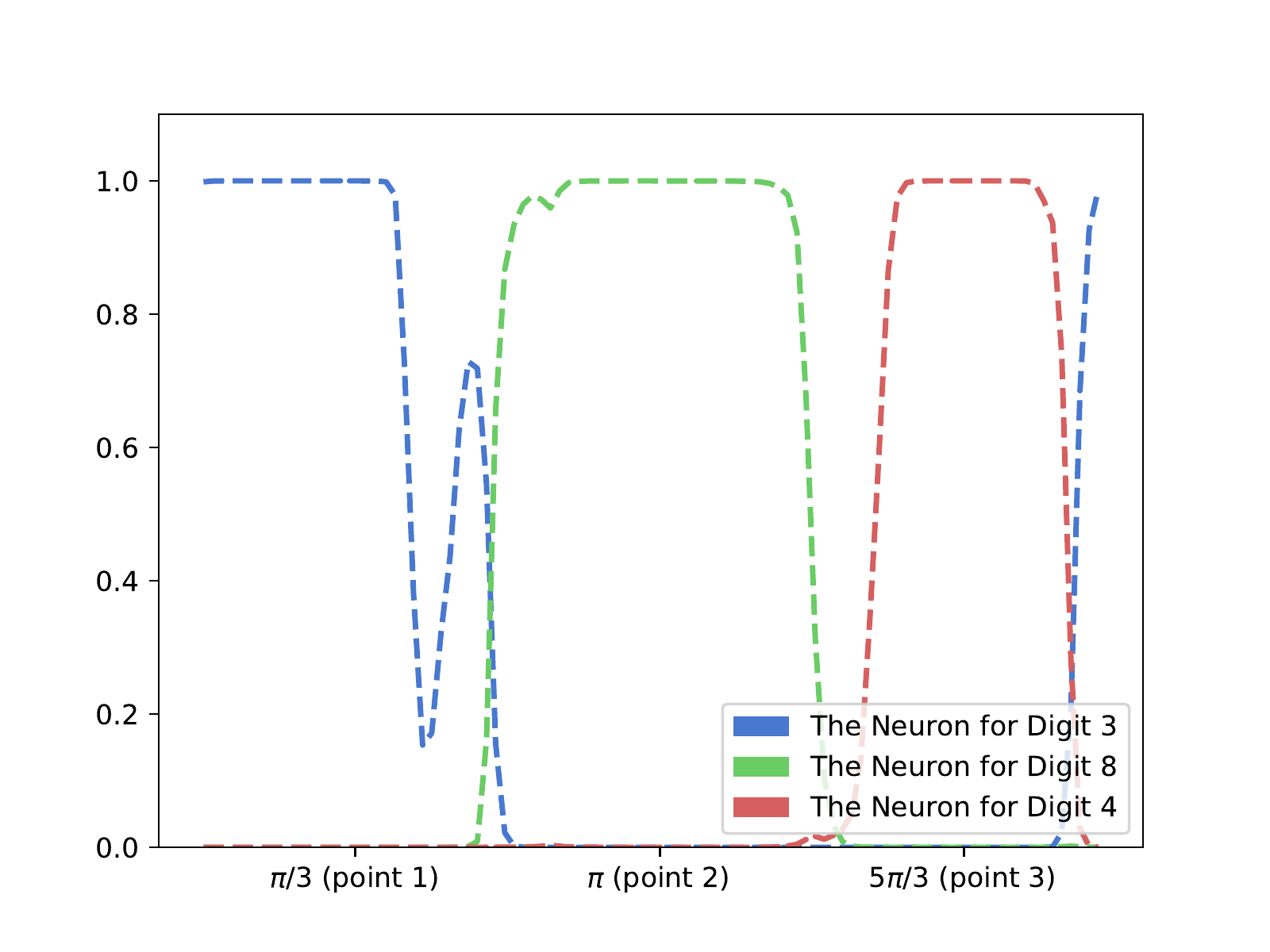}
	\includegraphics[width=0.24\textwidth]{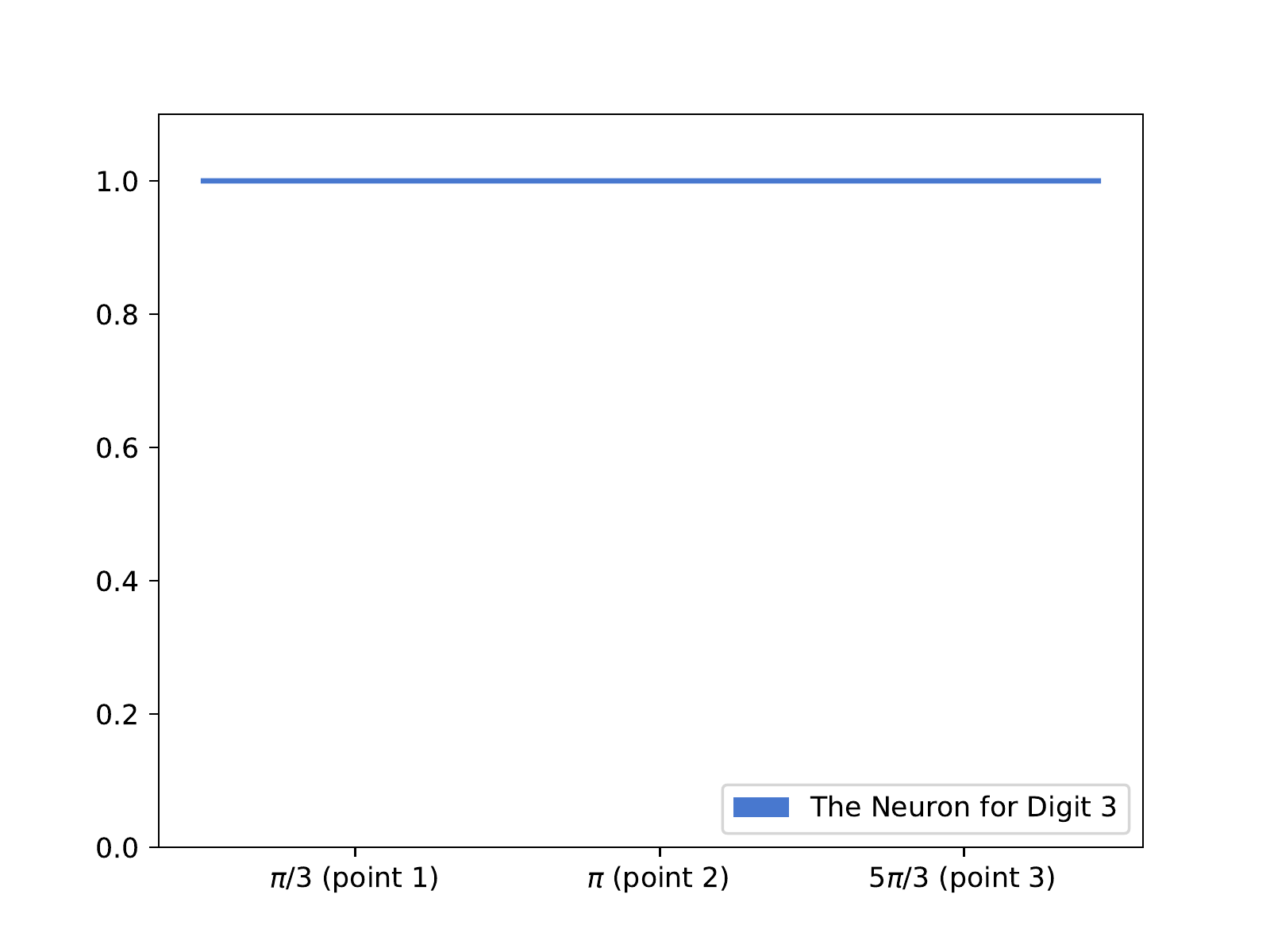}
	\includegraphics[width=0.24\textwidth]{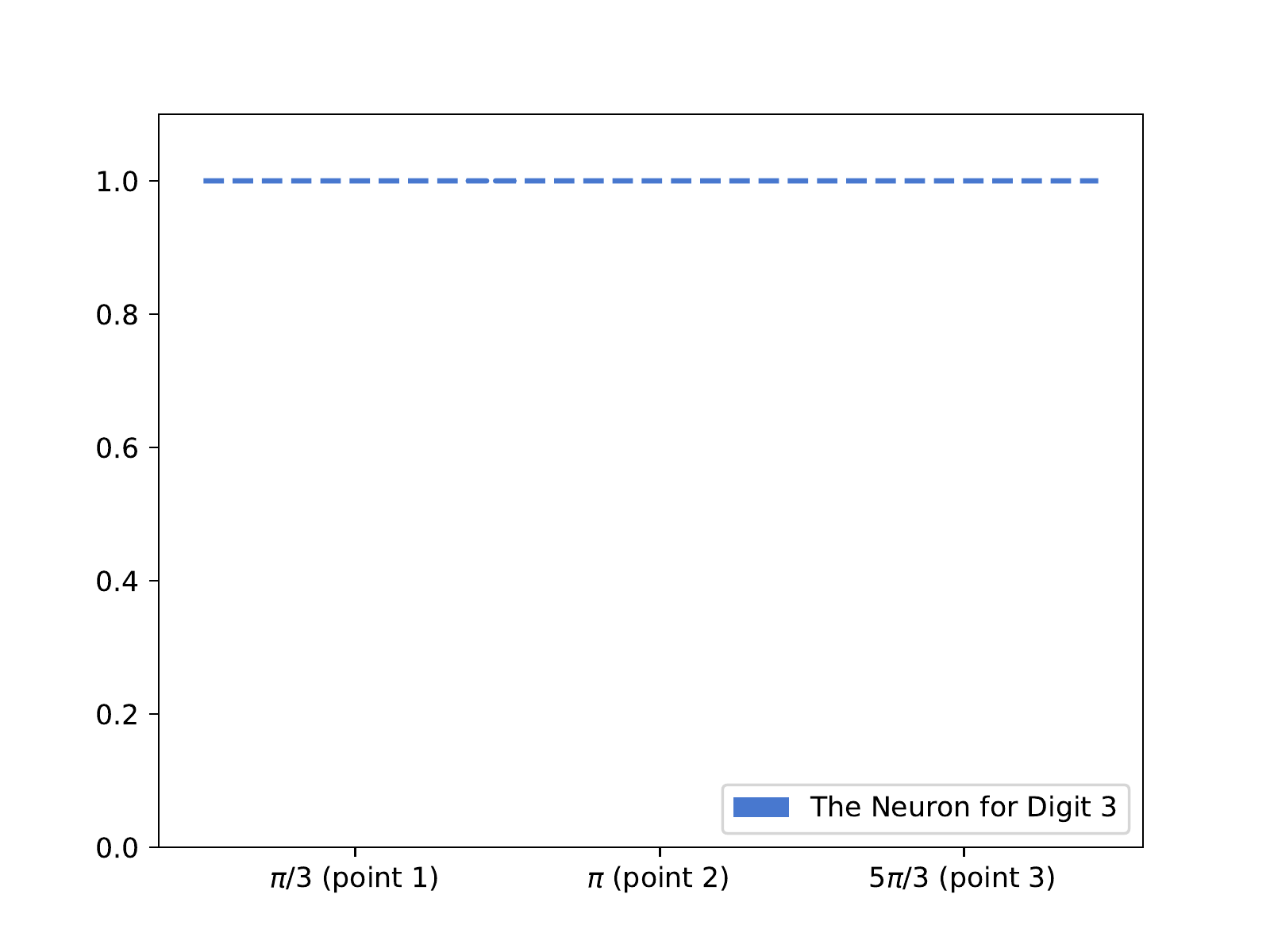}
	
	\includegraphics[width=0.24\textwidth]{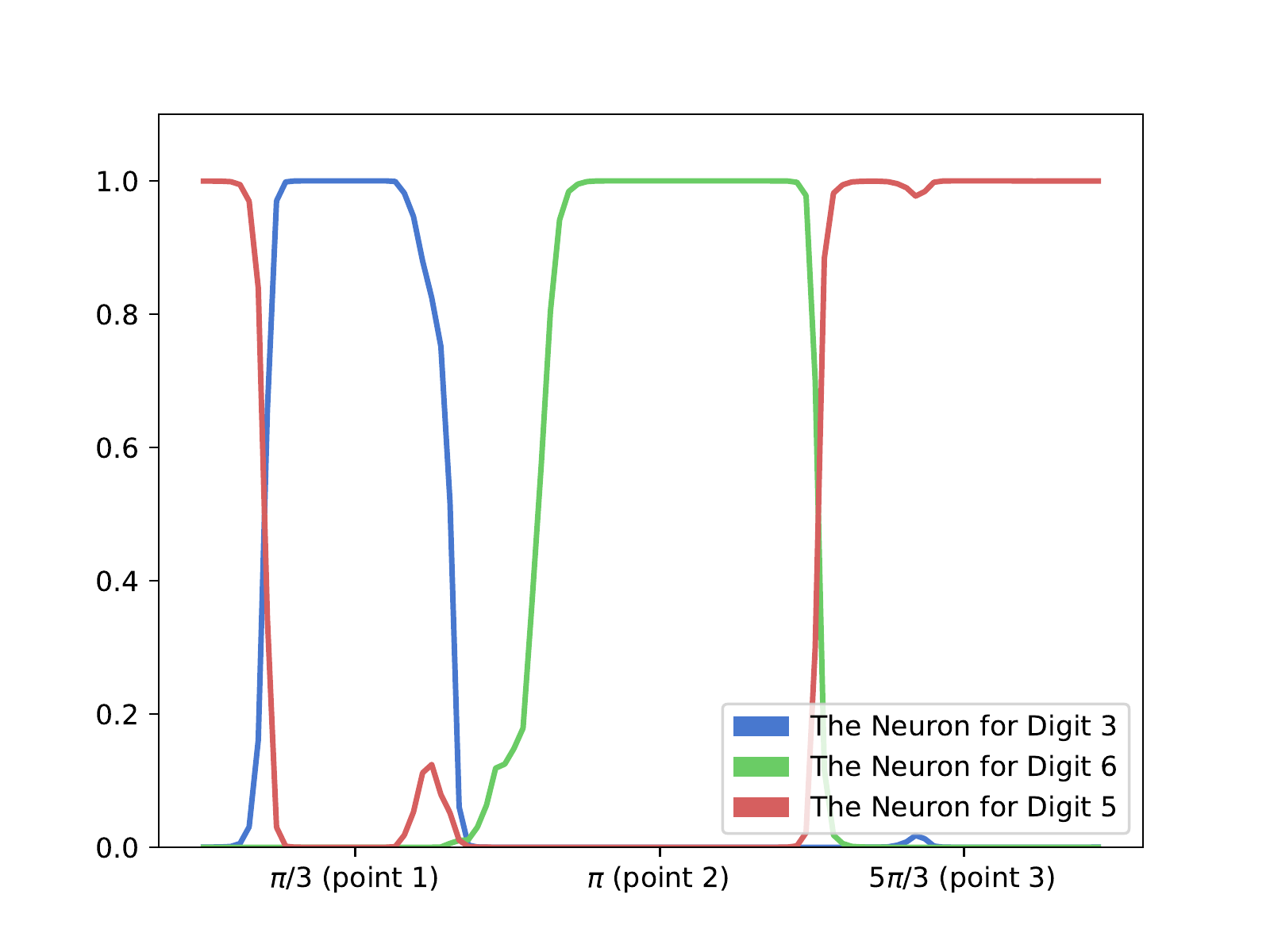}
	\includegraphics[width=0.24\textwidth]{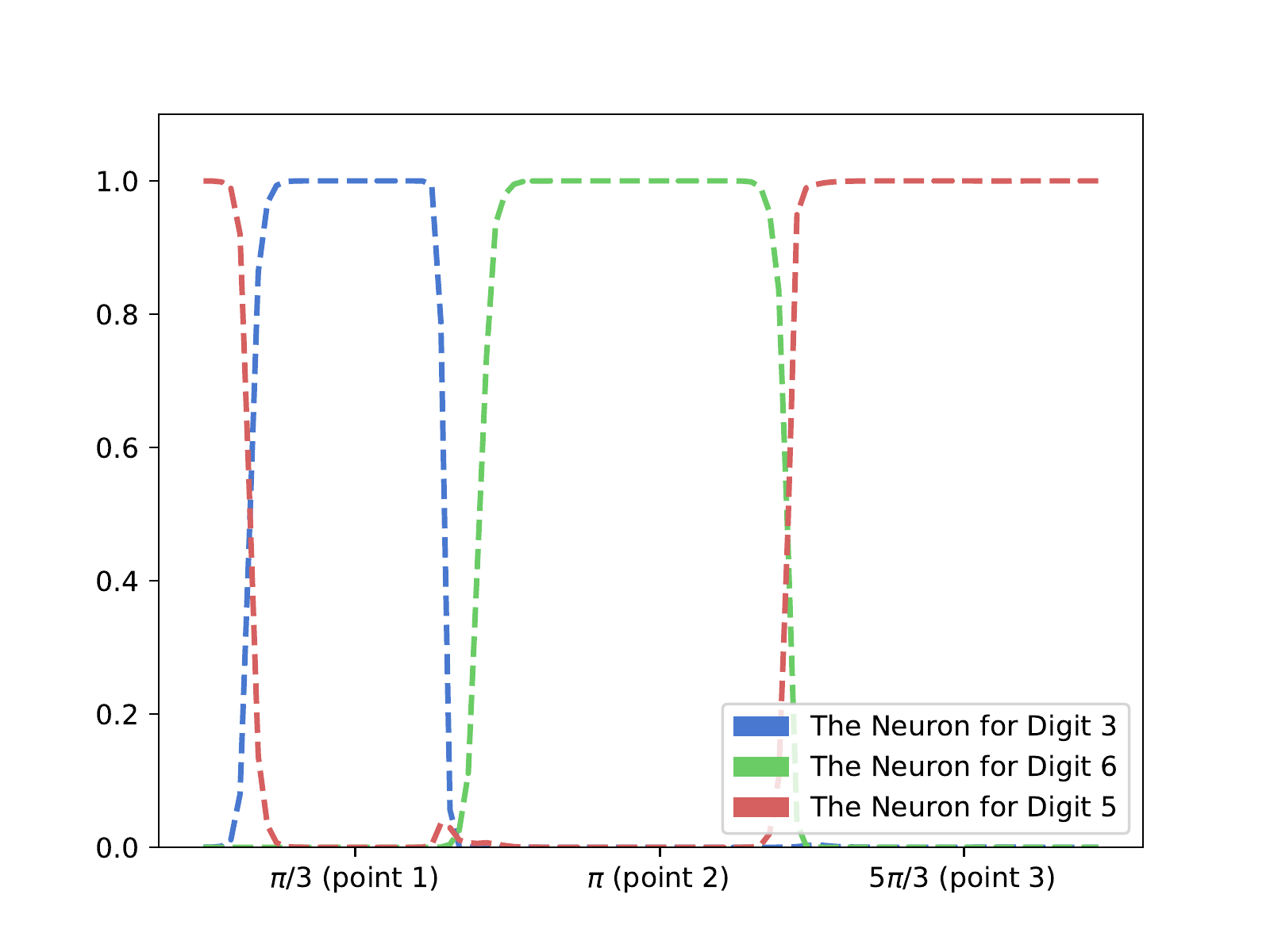}
	\includegraphics[width=0.24\textwidth]{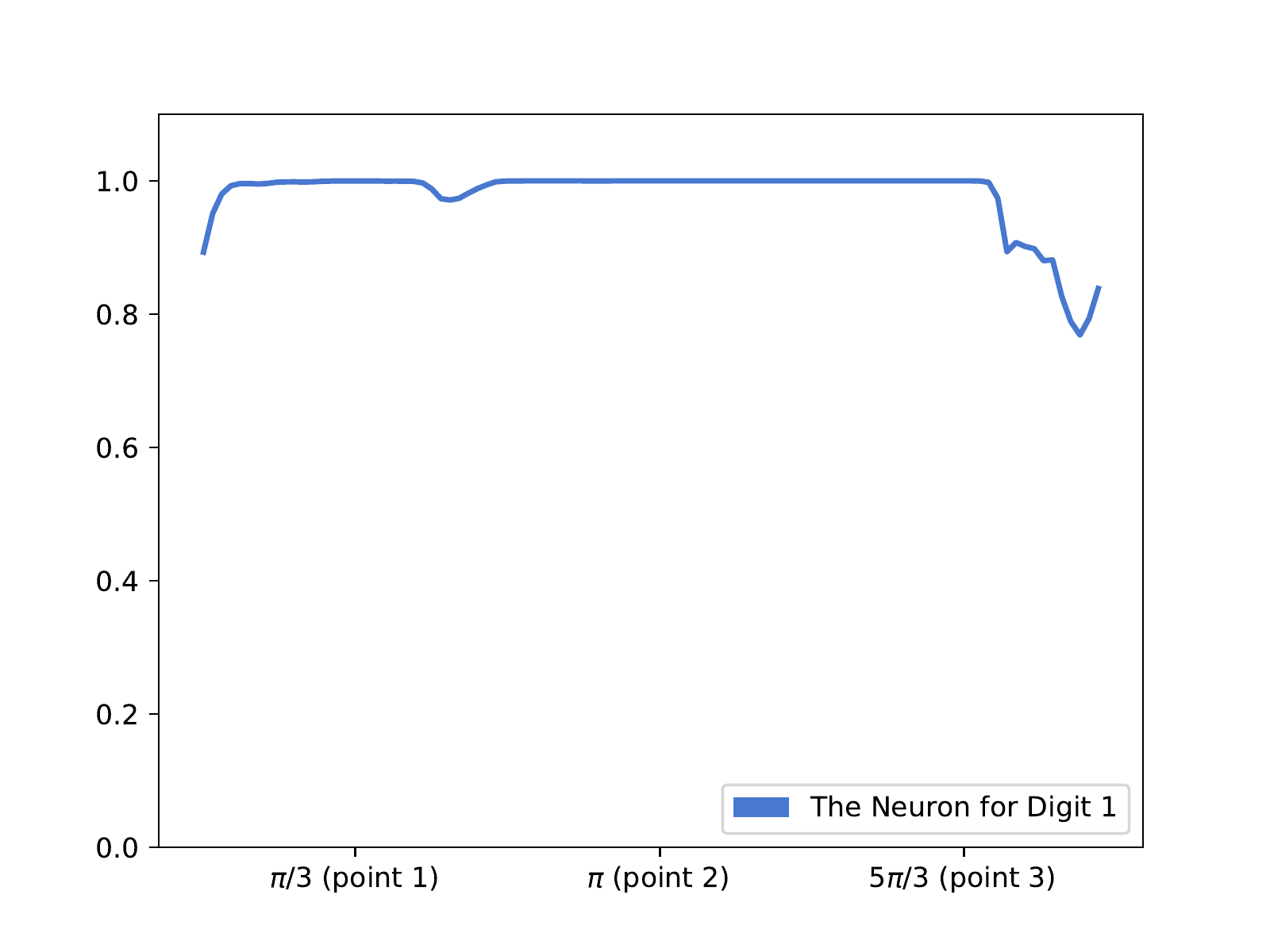}
	\includegraphics[width=0.24\textwidth]{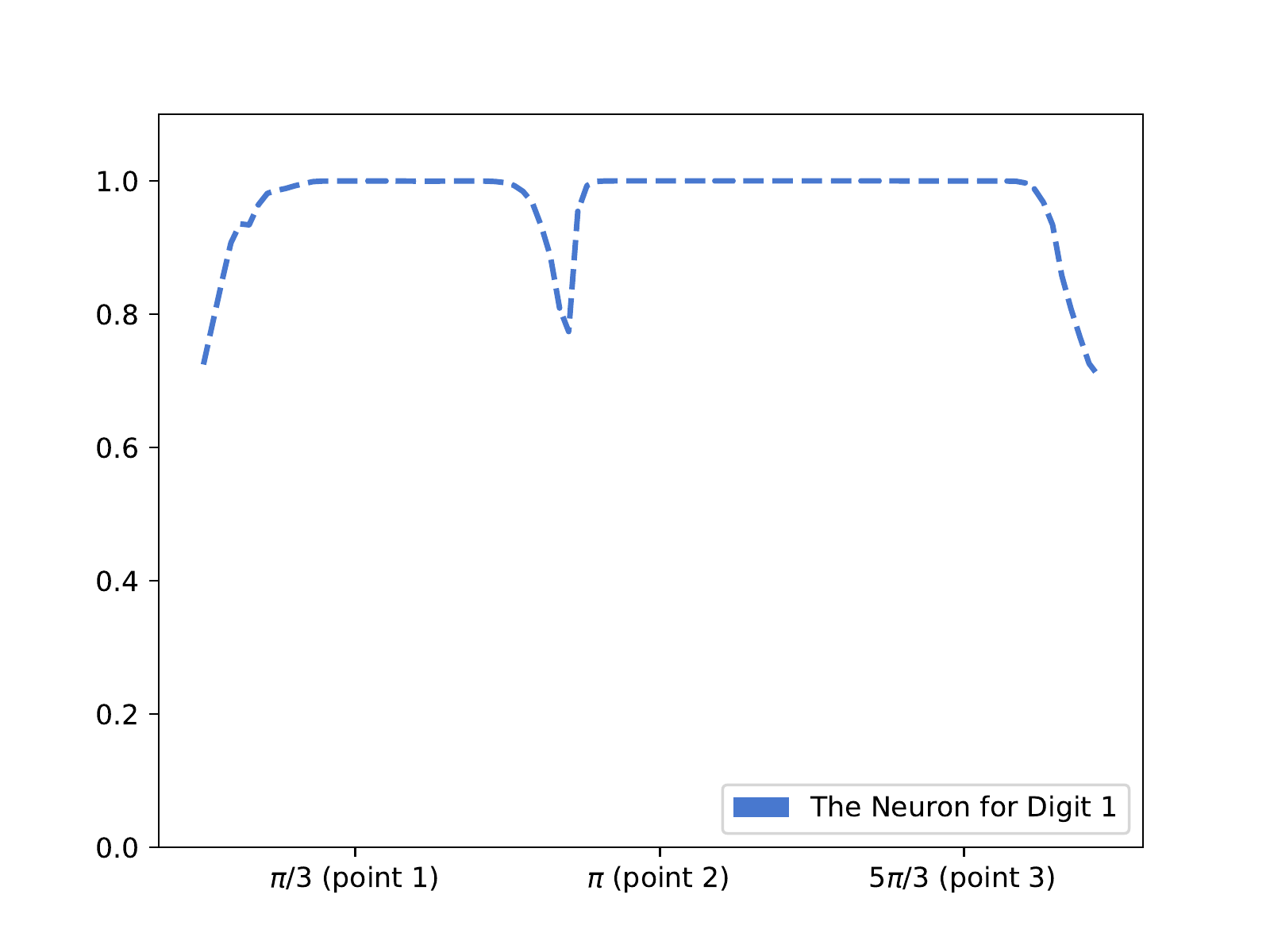}
	
	\includegraphics[width=0.24\textwidth]{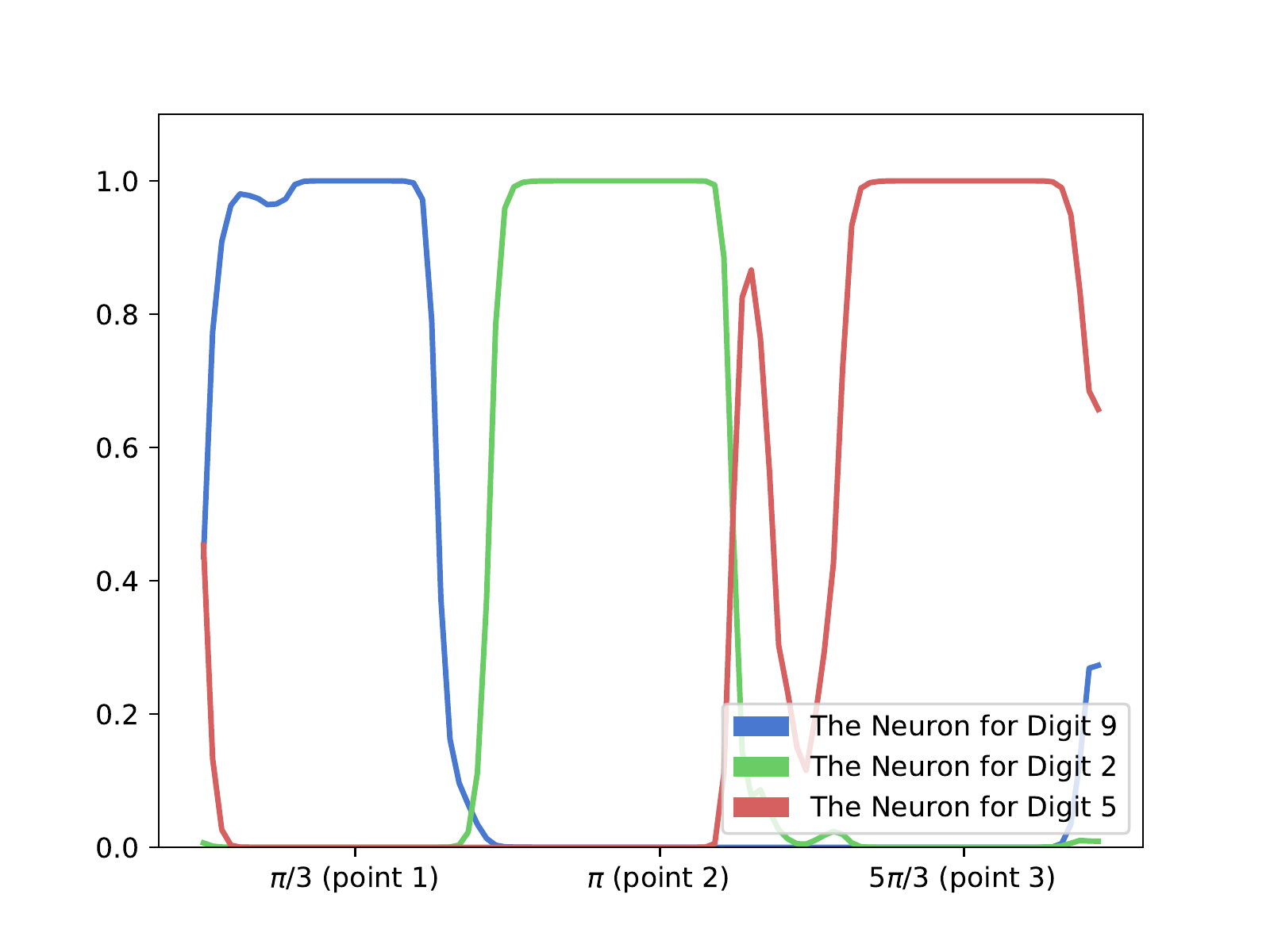}
	\includegraphics[width=0.24\textwidth]{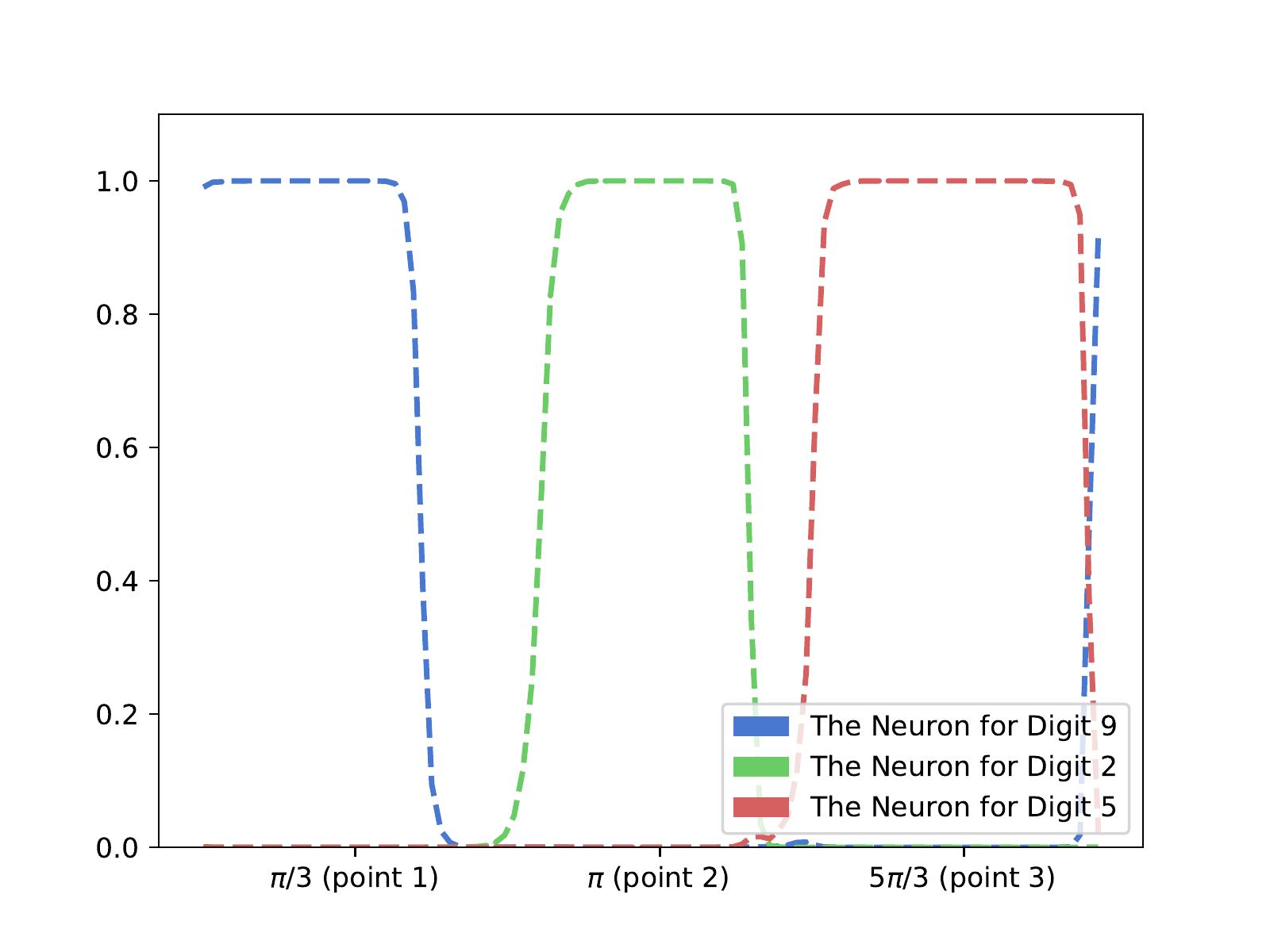}
	\includegraphics[width=0.24\textwidth]{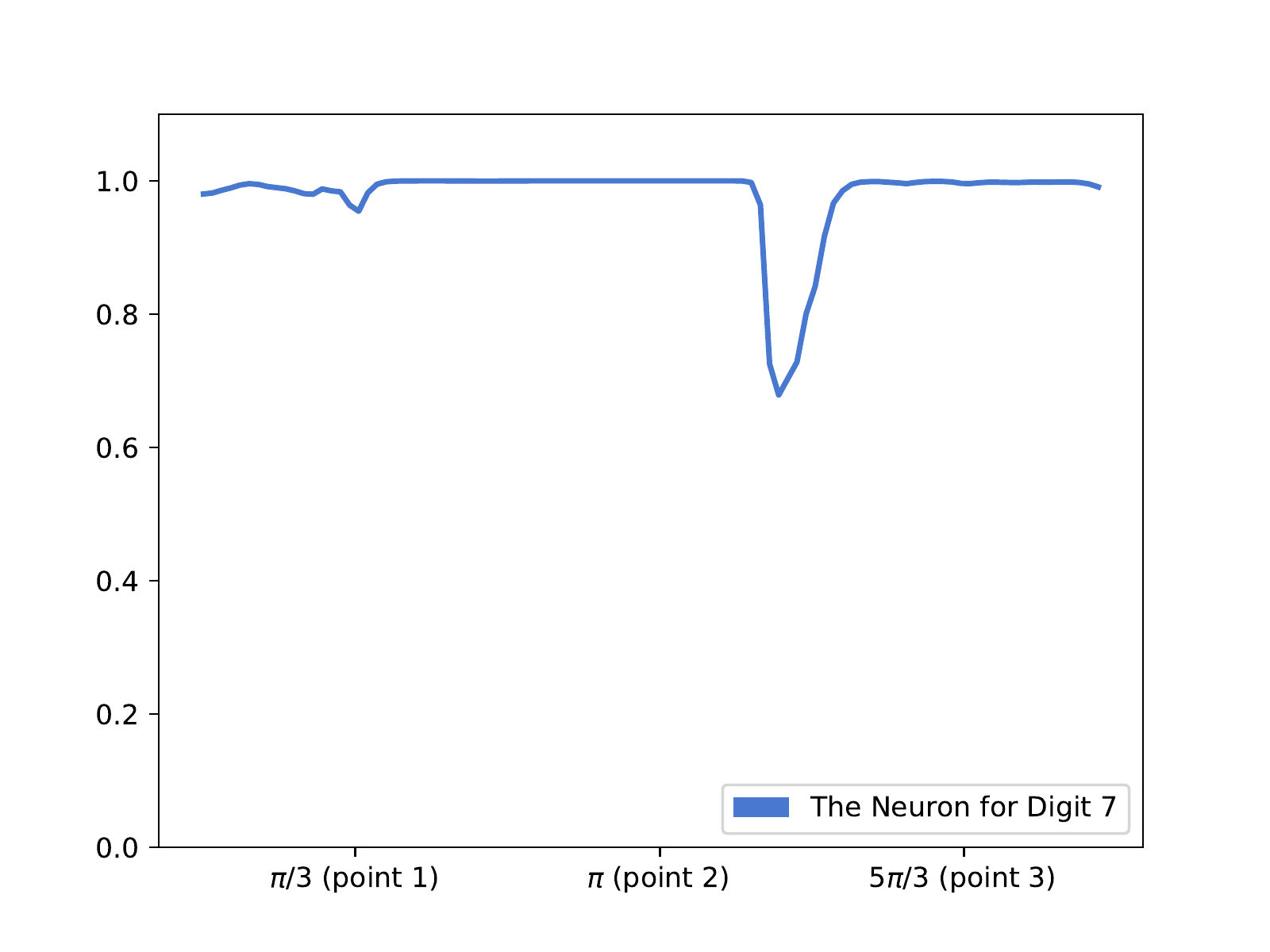}
	\includegraphics[width=0.24\textwidth]{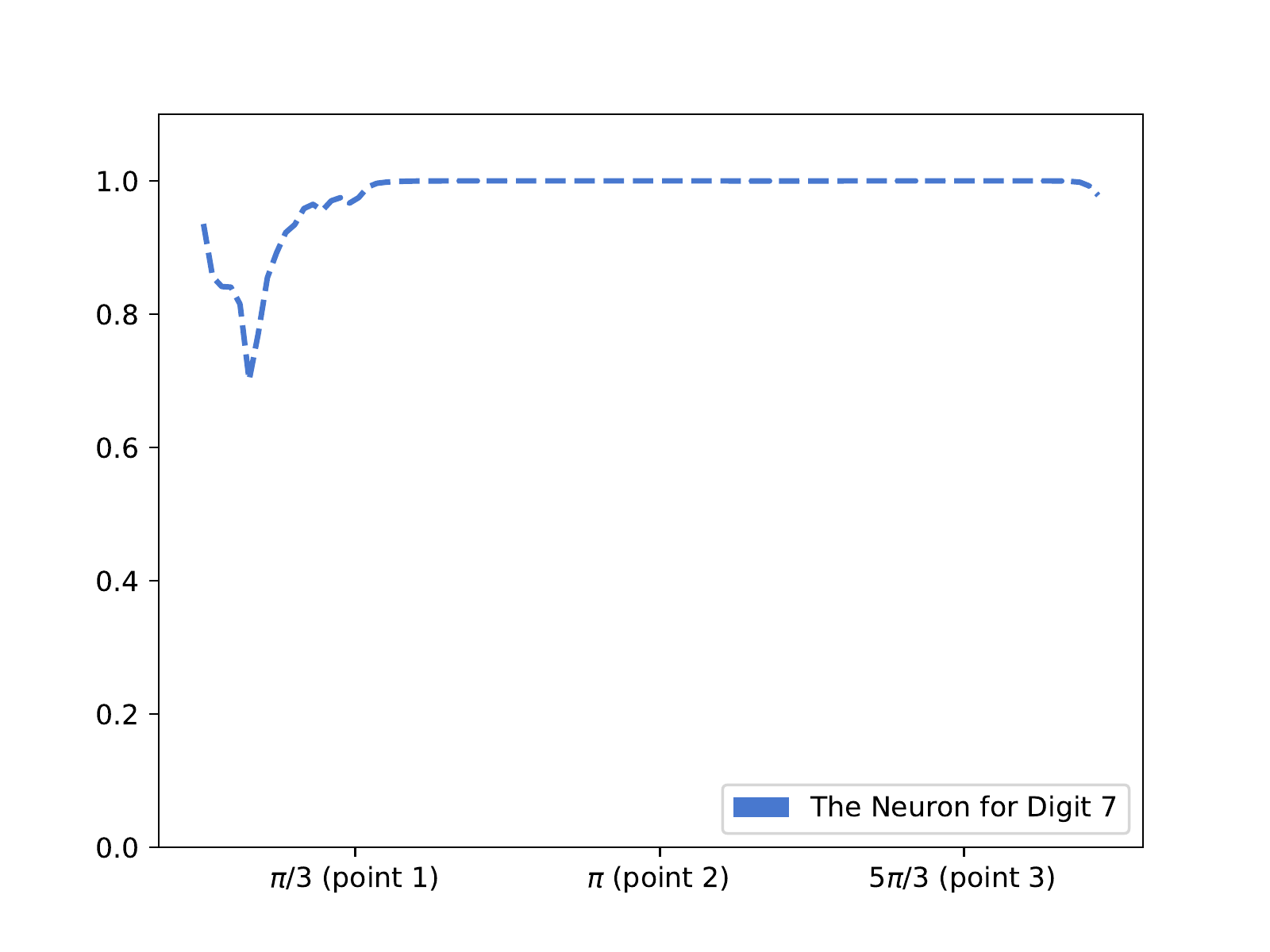}
	\includegraphics[width=0.24\textwidth]{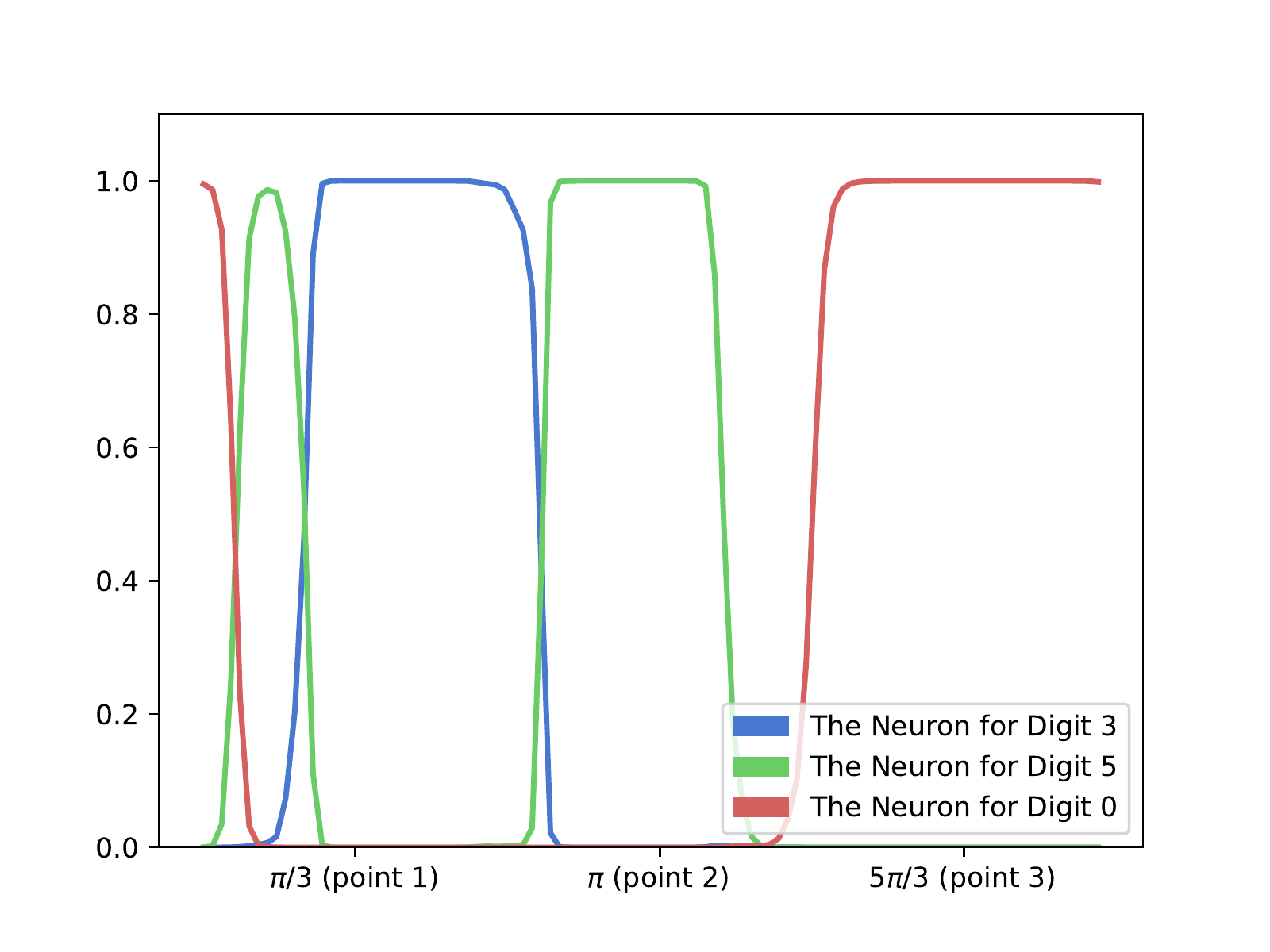}
	\includegraphics[width=0.24\textwidth]{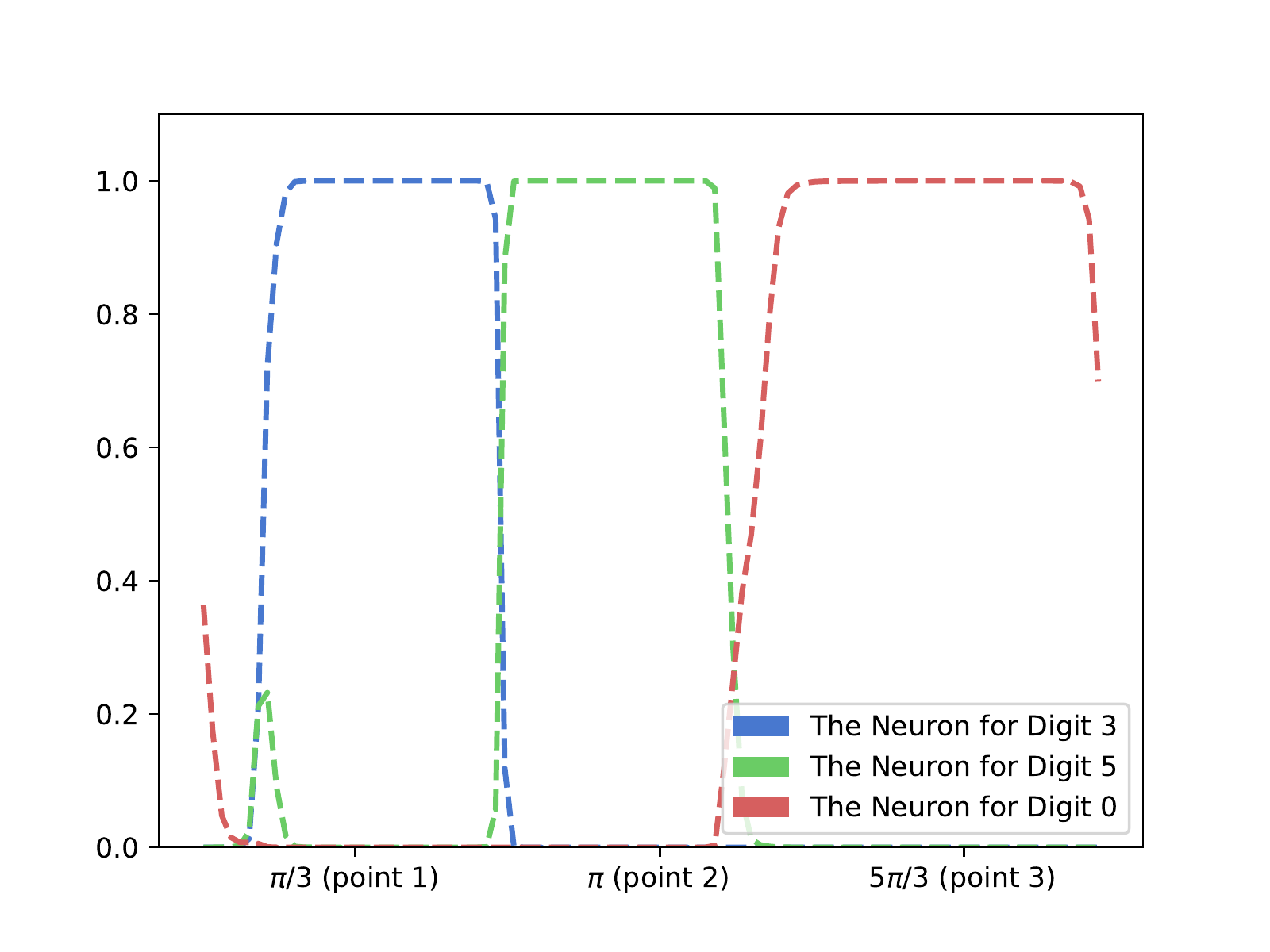}
	\includegraphics[width=0.24\textwidth]{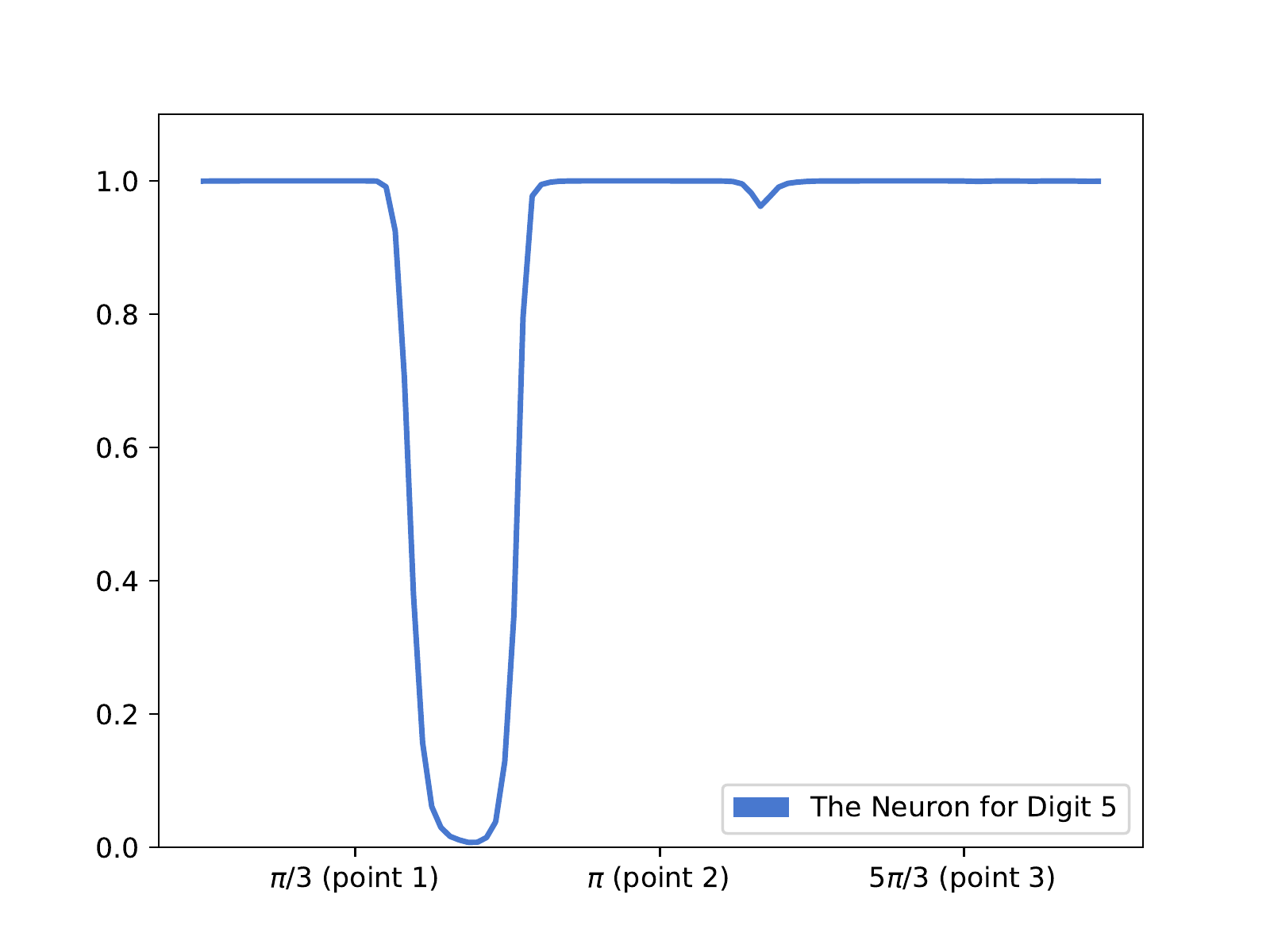}
	\includegraphics[width=0.24\textwidth]{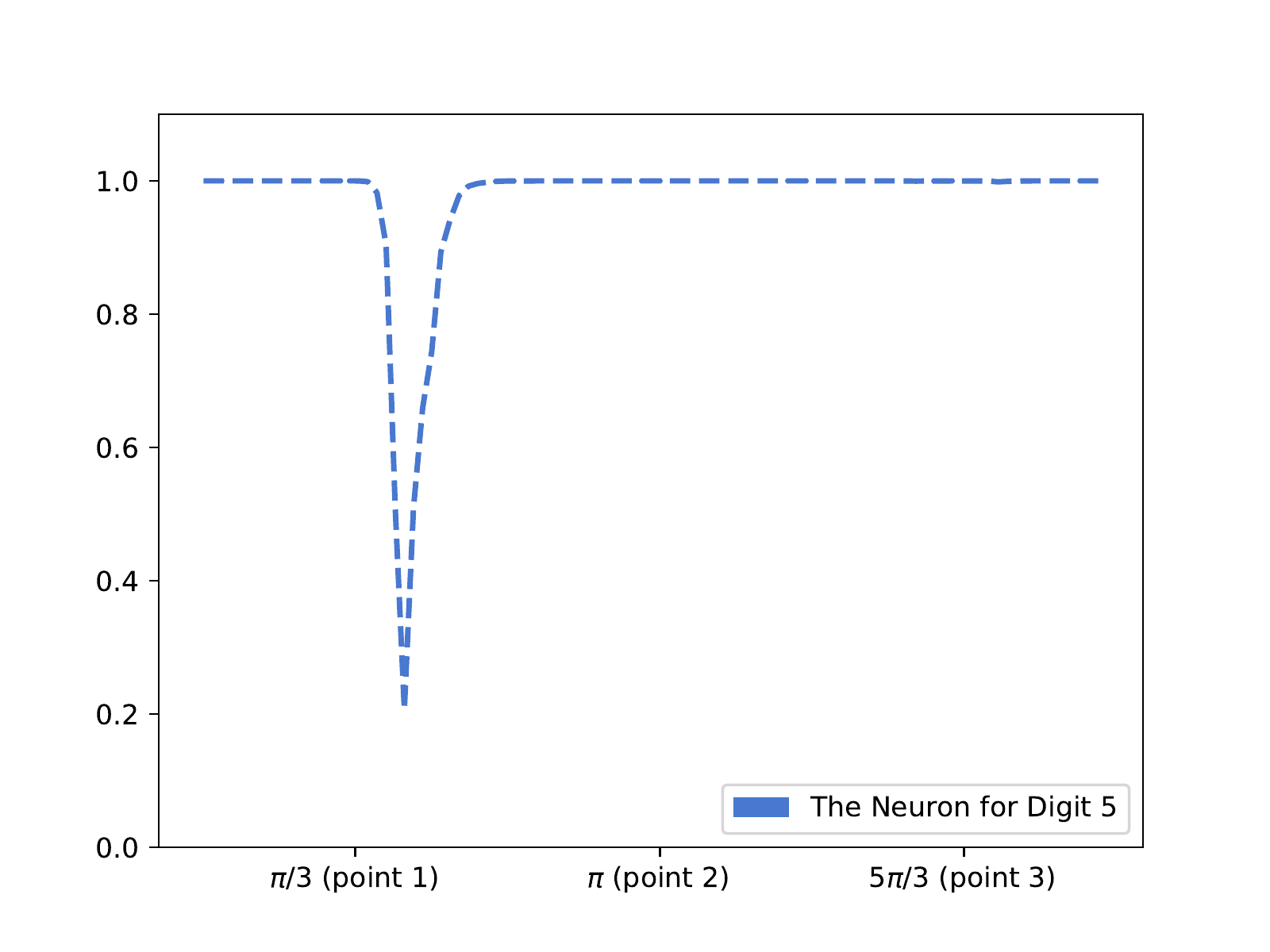}
	\subfigure[{\centering Vanilla CNNs; Trajectories through samples belonging to different classes.}]{
		\includegraphics[width=0.238\textwidth]{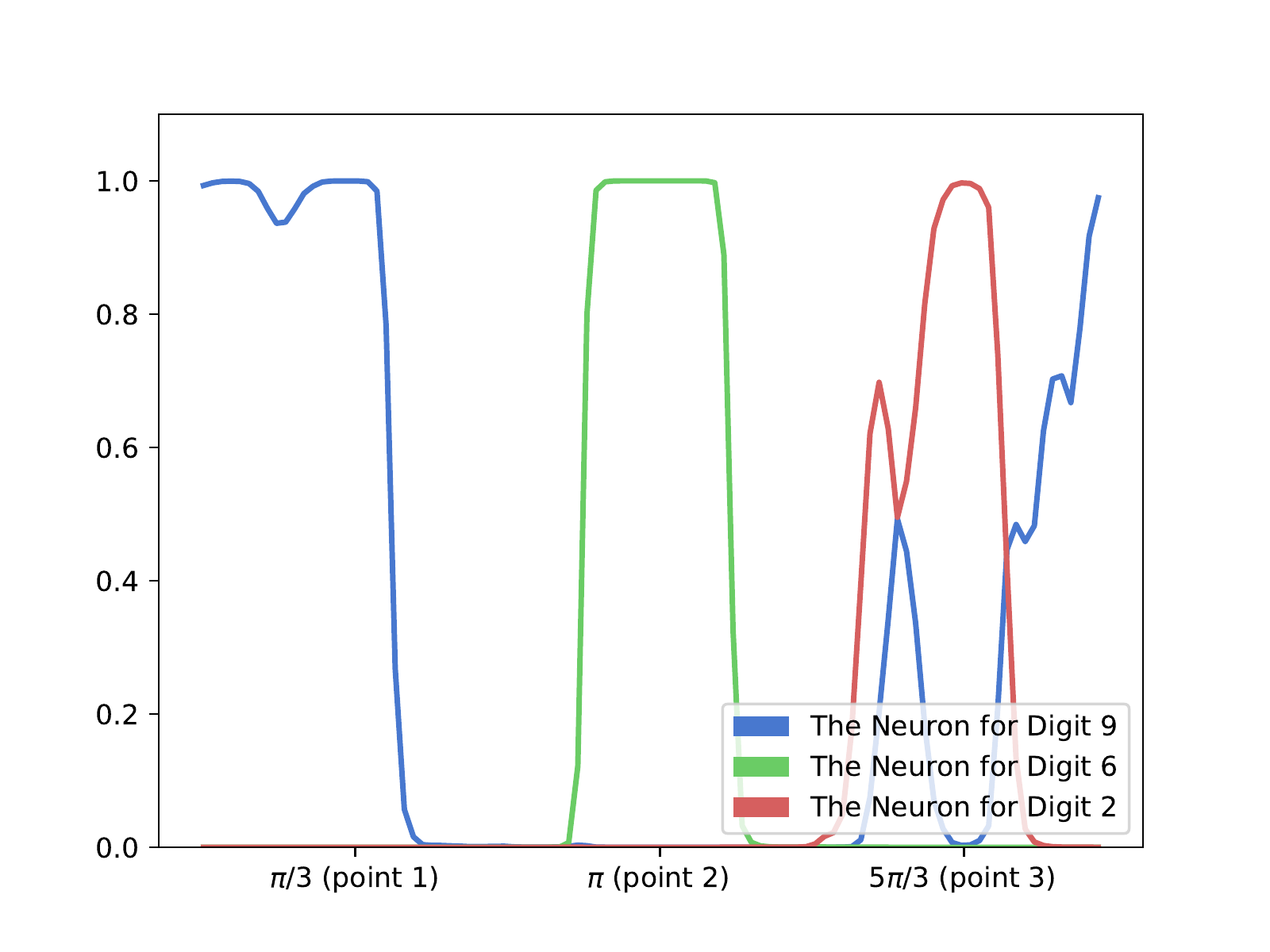}\label{fig:softmax:vanilla_d}}
	\subfigure[\centering AM CNNs; Trajectories through samples belonging to different classes.]{
		\includegraphics[width=0.238\textwidth]{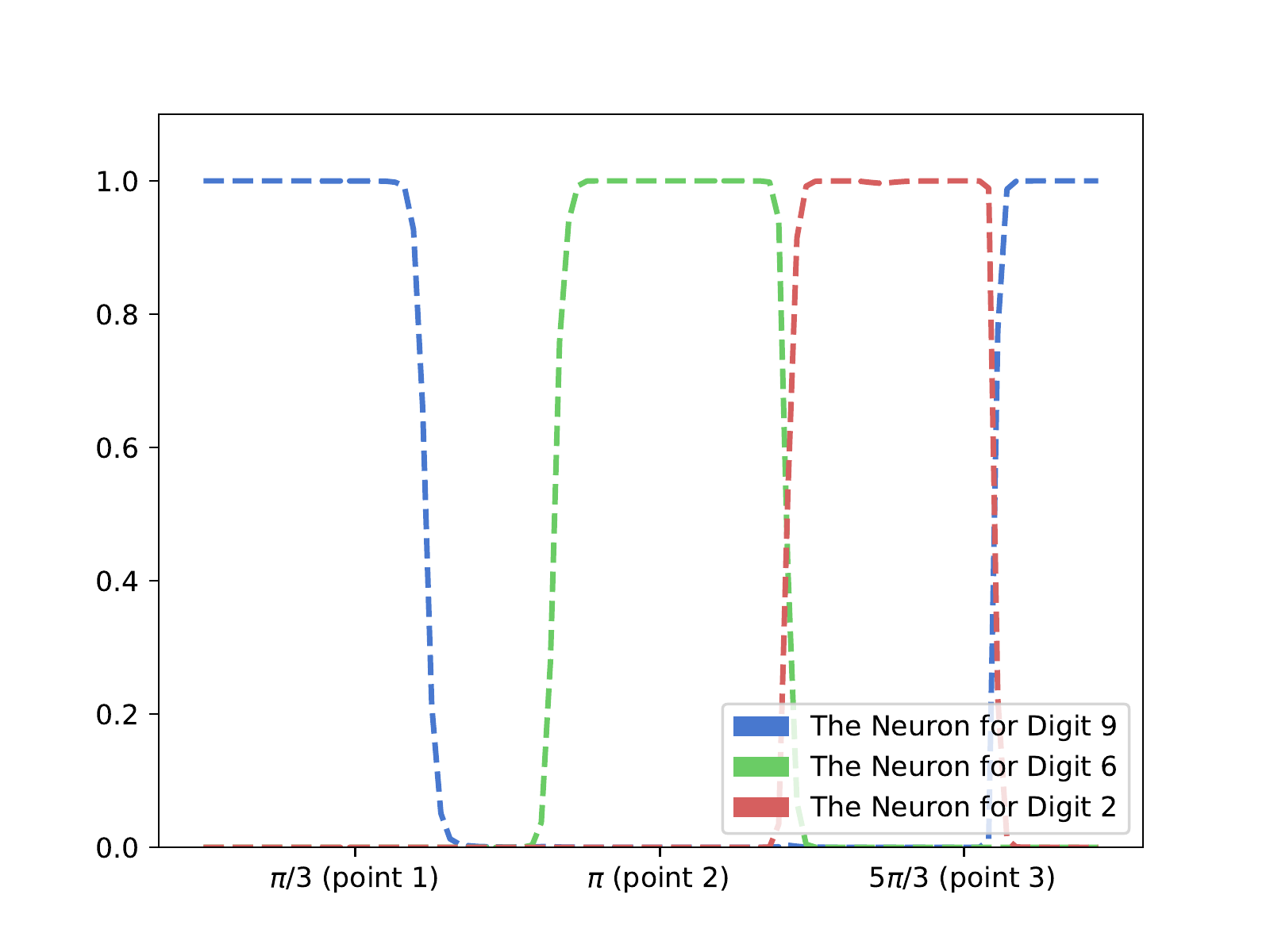}\label{fig:softmax:AM_d}}
	\subfigure[{\centering Vanilla CNNs; Trajectories through samples belonging to same classes.}]{
		\includegraphics[width=0.238\textwidth]{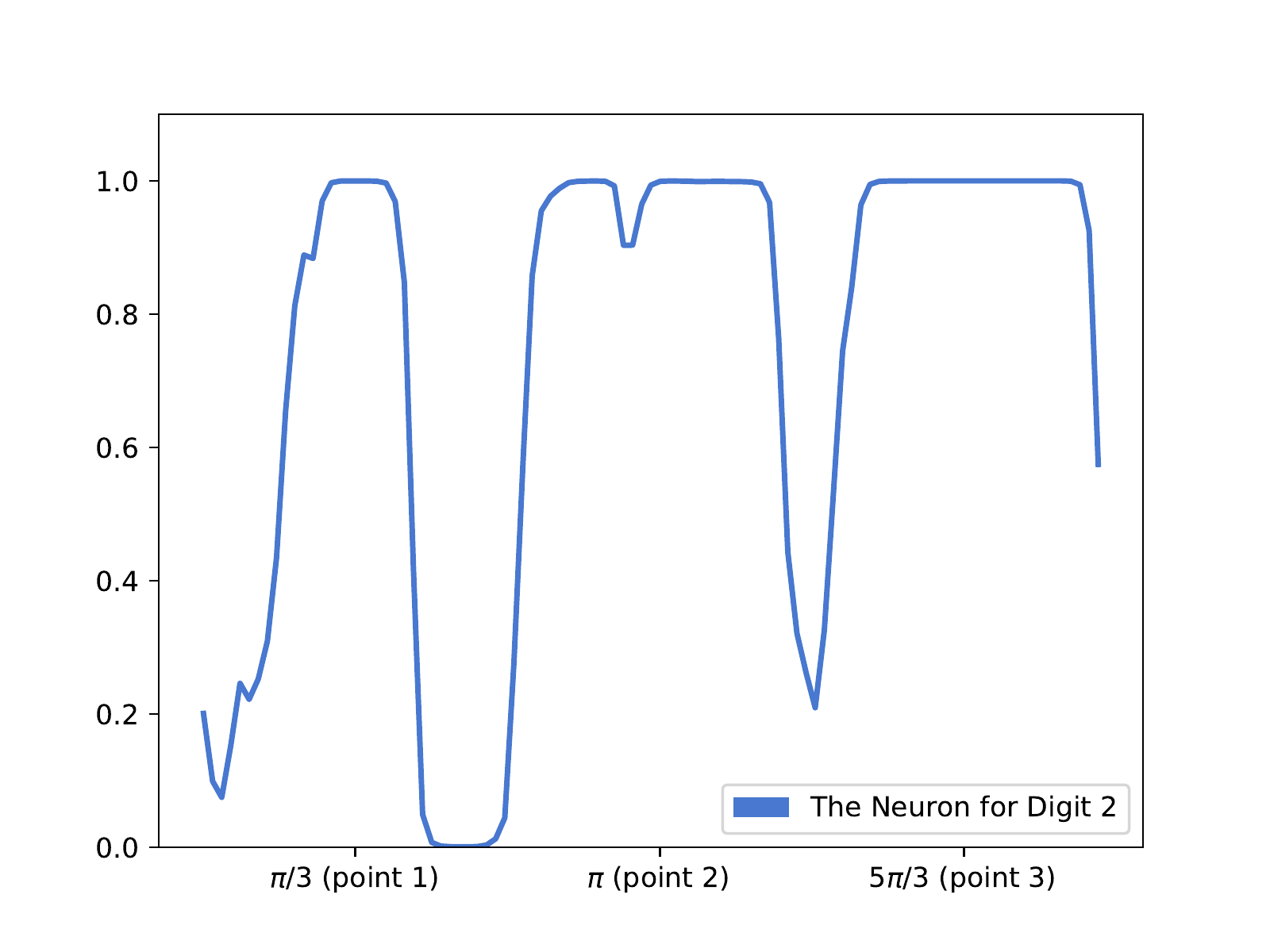}\label{fig:softmax:vanilla_s}}
	\subfigure[\centering AM CNN; Trajectories through samples belonging to same classes.]{
		\includegraphics[width=0.238\textwidth]{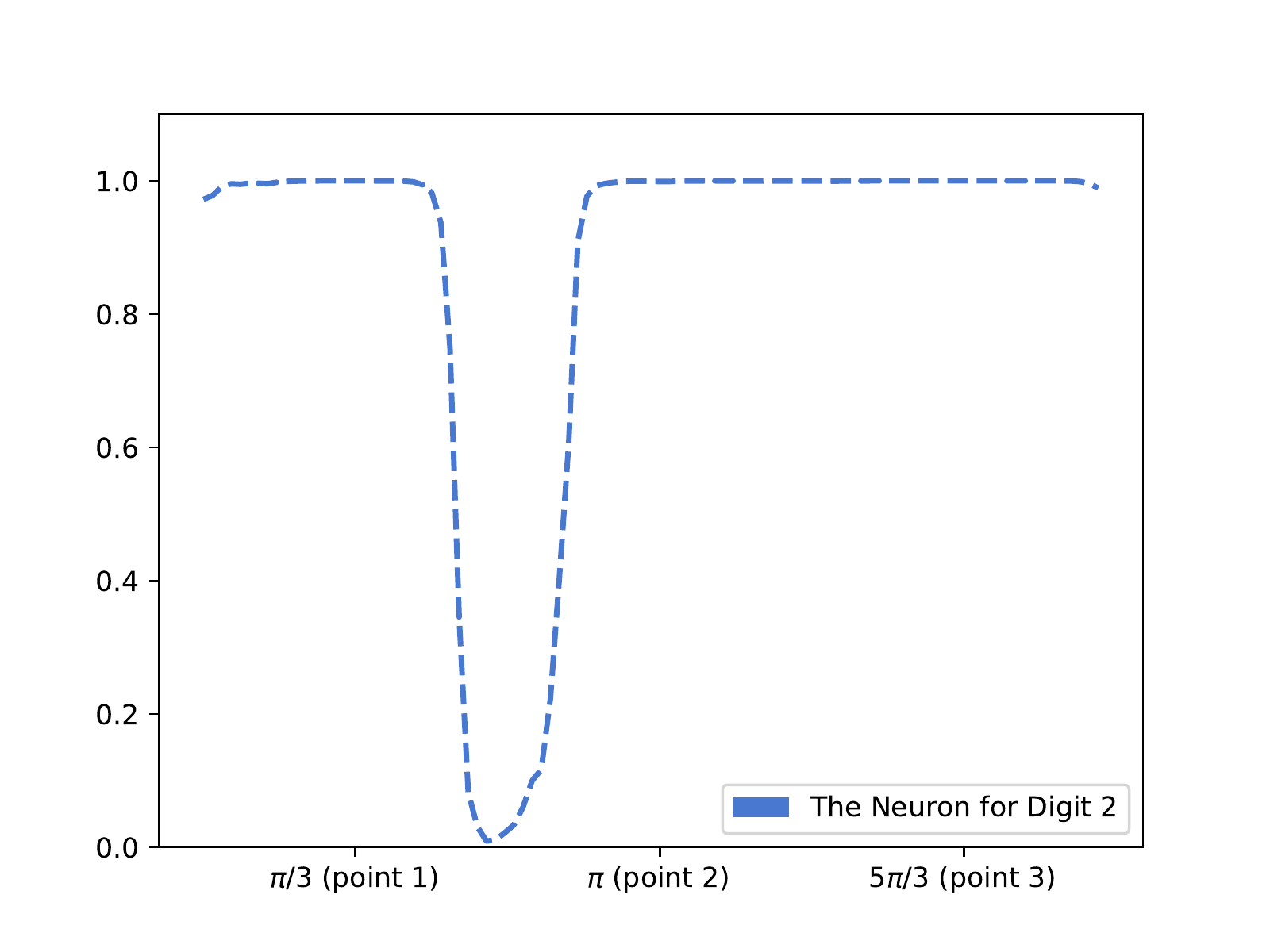}\label{fig:softmax:AM_s}}
	\caption{Figure (a) and Figure (b) depicted
		five visualization results for neurons along the cross-class ellipses.
		Compared to vanilla CNNs, AM CNNs seemed to be less sensitivity to the perturbations
		containing the information of samples belonging to different class.
		They yielded flatter
		curves around the real data points. Moreover, while vanilla CNNs
		demonstrated complex and disordered curves in-between real samples,
		AM CNNs demonstrated smooth, simple and almost linear curves.
		Figure (c) and Figure (d) depicted the five visualization results
		for neurons along the single-class ellipses, which were much more stable compared to that along
		cross-class ellipses. Nevertheless, similar to 
		Figure (a) and Figure (b), AM CNNs exhibited more stable curves and
		were less sensitivity to perturbations.}
	\label{fig:softmax} 
\end{figure*}

The visualization result was shown in Figure \ref{fig:softmax}. 
Figure \ref{fig:softmax:vanilla_d} and Figure \ref{fig:softmax:AM_d} depicted
five visualization results for neurons along the cross-class ellipses. Both vanilla CNNs
and AM CNNs yielded the correct predictions for
the picked real data points, i.e., the points representing the picked test samples in the ellipses.
Their neurons yielded the highest activation at the real data points belonging to corresponding classes.
Nevertheless, compared to vanilla CNNs, AM CNNs seemed to be less sensitivity to the perturbations
containing the information of samples belonging to different class.
They yielded flatter
curves around the real data points. Moreover, while vanilla CNNs
demonstrated complex and disordered curves in-between real samples,
AM CNNs demonstrated smooth, simple and almost linear curves.

Figure \ref{fig:softmax:vanilla_s} and Figure \ref{fig:softmax:AM_s} depicted the five visualization results
for neurons along the single-class ellipses, which were much more stable compared to that along
cross-class ellipses. Nevertheless, similar to 
Figure \ref{fig:softmax:vanilla_d} and Figure  \ref{fig:softmax:AM_d}, AM CNNs exhibited more stable curves and
were less sensitivity to perturbations.

From the visualization results in Figure \ref{fig:softmax}, we found out that compared to vanilla CNNs,
AM CNNs learned smoother and more stable
curves along single-class and cross-class ellipses. It made AM CNNs less sensitivity to
the perturbations containing the information of samples belonging to the same or different classes.
Based on the visualization result, however, we conjectured that AM CNNs not only learned smoother and more stable curves
along single-class and cross-class ellipses, but all their learned landscapes. It made AM CNNs less sensitivity
to all kinds of perturbations, which brought about high generalization abilities of CNNs \cite{9412496,novak2018sensitivity}. 
To verify the conjecture, we compared the sensitivity scores between AM CNNs and vanilla CNNs in Section \ref{sec:sensitivity}.

\subsubsection{Sensitivity and Generalization}
\label{sec:sensitivity}
\begin{table}[ht]
	\centering
	\caption{The sensitivity scores of AM equipped and vanilla CNNs.
		Compared to vanilla CNNs, AM CNNs yielded smaller sensitivity
		scores for both $\mathbf{S}$ \cite{9412496} and $\mathbf{J}$ \cite{novak2018sensitivity} on
		all the listed datasets.
	}
	\renewcommand\arraystretch{1.2}
	
		\subtable[Model sensitivities measured by sensitivity metric $\mathbf{J}$]{
		\centering
		\begin{tabular}{c|cccc}
			\hline 
			Datasets &MNIST &FashionMNIST&CIFAR-10\\ 
			\hline
			Vanilla CNNs&677&11437&17347\\
			AM CNNs&357&7492&12408\\
			\hline	
	\end{tabular}}
	\subtable[Model sensitivities measured by sensitivity metric $\mathbf{S}$]{
		\centering
		\begin{tabular}{c|cccc}
			\hline 
			Datasets &MNIST &FashionMNIST&CIFAR-10\\ 
			\hline
			Vanilla CNNs&7.72&93.10&974.40\\
			AM CNNs&2.61&47.04&235.54\\
			\hline	
	\end{tabular}}
	\label{tab:sen}
\end{table}

Compared to vanilla CNNs, AM equipped CNNs learned smoother and more stable curves along
single-class or cross-class ellipses, which brought about their less sensitivity
to perturbations containing information of samples belonging to the same or different classes.
This section compared two kinds of sensitivity scores,
$\mathbf{S}$ \cite{9412496} and $\mathbf{J}$ \cite{novak2018sensitivity},
between vanilla CNNs and AM CNNs, which represented their sensitivity to all kinds of perturbations.
It would demonstrate that with smoother and more stable
learned landscapes, AM CNNs yielded smaller sensitivity scores. It was proved \cite{9412496,novak2018sensitivity}
to made them generate better.

The sensitivity result was listed in Table \ref{tab:sen}. 
Compared to vanilla CNNs, AM CNNs yielded smaller sensitivity
scores for both $\mathbf{S}$ \cite{9412496} and $\mathbf{J}$ \cite{novak2018sensitivity} on
all the listed datasets. 
\uline{
	The key for AM improving the performance of CNNs was the feature map
multiplication operator bringing about smoother and more stable learned landscapes, which made
them generalize better. 
Section \ref{sec:do} explored the role AM/feature map multiplication played
on the learned landscapes, which was conjectured to be a regularization role.
}

\subsection{Does Attention Mechanism Play a Regularization Role in CNNs?}
\label{sec:do}
\subsubsection{Overview}
Section \ref{sec:visualization} illustrated that
compared to vanilla CNNs, AM/feature map multiplication
equipped CNNs learned smoother and more stable curves along single-label or cross-label ellipses,
without any external supervisors.
The smoother and more stable landscapes made them generalize better to
test datasets. 
We conjectured that AM played a regularization role
on the learned landscapes of CNNs and verified the regularization role of AM
by using it in conjunction with the augmentation method, mixup \cite{zhang2018mixup}.

Mixup augmented the training dataset by combining pairs of input samples and their labels, formally,
\begin{equation*}
\begin{split}
&\tilde{x} = \lambda x_i + (1 - \lambda) x_j\\
&\tilde{y} = \lambda y_i + (1 - \lambda) y_j,
\end{split}
\end{equation*}
where $(x_i, y_i)$ and $(x_j, y_j)$ were two samples drawn at random
from training data, and $\lambda \in [0,1]$ \cite{zhang2018mixup}.
As shown in the equation, when the two samples belonged to the same/different classes, mixup
regularized the learned curve of CNNs along single-class/cross-class ellipses with a supervised manner.
Thus, mixup might have the same effect on the learned landscapes compared with AM.
To compare the role mixup and AM played on CNNs, two kinds of experiments were conducted.

\begin{itemize}
\item The performances between mixup and AM equipped CNNs were compared.

\item The performance of mixup equipped CNNs was compared
to that of CNNs equipped with both mixup and AM.
\end{itemize}

We verified that both mixup and AM improved the performance of CNNs by regularizing
their learned landscapes. Thus, the effectiveness of AM might owe to the
regularization role it played.

\subsubsection{Implementation}

To verify the role AM played on CNNs, four kinds of models were trained and tested,
CNNs (vanilla CNNs),
CNNs$_{Aug}$ (mixup equipped CNNs),
AM CNNs (attention mechanism equipped CNNs),
and AM CNNs$_{Aug}$ (both mixup and attention mechanism equipped CNNs).
For vanilla CNNs, we built toy CNNs with three convolutional layers and a fully-connected layer.
The number of kernels in each convolutional layer was 32, 64, and 64 respectively.
SE modules were embedded into the vanilla CNNs for 
AM CNNs and AM CNNs$_{Aug}$.
All the models were trained on MNIST and CIFAR-10 datasets 
using momentum optimizer for 200 epochs with a batch size of 400
and a momentum of 0.9.

\subsubsection{Result}
\begin{table}[ht]
	\centering
	\caption{The result verifying the role AM played in CNNs.
		It was interesting to see, mixup did not improve the performance of AM equipped CNNs.
		AM CNNs$_{Aug}$ yielded a comparable performance to AM CNNs, i.e., 99.09\% and 99.03\% on MNIST dataset;
		79.64\% and 79.89\% on CIFAR-10 dataset.
		It indicated that with a smooth and stable landscape learned by AM equipped CNNs,
		they did not need the augmentation method which regularizing the landscapes learned by CNNs in a 
		supervised manner. AM played a mixup-liked role on CNNs which was of course to be
		a regularization role on the landscapes learned by CNNs.
	}
	\renewcommand\arraystretch{1.2}
	\label{tab:mixup}
	\begin{tabular}{c|cccc}
		\hline 
		Datasets &CNNs & CNNs$_{Aug}$ & AM CNNs&AM CNNs$_{Aug}$\\
		\hline
		MNIST&98.61&99.02&99.03&99.09\\
		CIFAR-10&79.57&80.34&80.40&80.45\\
		\hline	
	\end{tabular}
\end{table}
The result verifying the role AM played in CNNs was shown in Table \ref{tab:mixup}.
As shown in the table, CNNs$_{Aug}$ and AM CNNs outperformed vanilla CNNs by the same margin,
and improved the performance over vanilla CNNs by 0.95\% and 0.99\% on CIFAR-10, respectively.
It was interesting to see, mixup did not improve the performance of AM equipped CNNs.
AM CNNs$_{Aug}$ yielded a comparable performance to AM CNNs, i.e., 99.09\% and 99.03\% on MNIST dataset;
79.64\% and 79.89\% on CIFAR-10 dataset.
It indicated that with a smooth and stable landscape learned by AM equipped CNNs,
they did not need the augmentation method which regularizing the landscapes learned by CNNs in a 
supervised manner. AM played a mixup-liked role on CNNs which was of course to be
a regularization role on the landscapes learned by CNNs.

\subsection{Conclusion}
This section gave an enlightenment for understanding AM in CNNs.
Firstly, section \ref{sec:key} found that the feature map multiplication was crucial
for the effectiveness of AM. Secondly, in section \ref{sec:landscape}, we visualized and 
compared the learned
curves between vanilla CNNs and AM equipped CNNs
along single-class and cross-class ellipses and discovered that the latter
learned smoother and more stable landscapes compared to vanilla CNNs.
The smooth and
stable landscapes brought about less sensitivity to the perturbations of input and good generation
ability, which was demonstrated to be the main reason for the effectiveness of AM.
Finally, in section \ref{sec:do}, we verified the regularization role of AM which
made the learned landscapes smooth and stable,
by using it in conjunction with the augmentation method, mixup.
It demonstrated that AM played a mixup-liked role on CNNs
which brought about simple behaviors in-between input samples,
i.e., smooth and stable curves along single-class and cross-class ellipses,
without any external supervisors.

Knowing the importance of feature map multiplication and the regularization role AM
played, we proposed a new CNN architecture FMMNet (Feature Map Multiplication Netwrok). 
It was the first CNN architecture motivated by the effectiveness of feature map multiplication, not
the conventional visual attention. The experiments of FMMNet would be illustrated in section \ref{sec:FMMNet}.

\section{Experiments of FMMNet}
\label{sec:FMMNet}

\subsubsection{Overview}
As illustrated in section \ref{sec:enlightenment}, feature map multiplication
bringing about high order non-linearity to CNNs, was crucial for
the effectiveness of AM. 
It made the landscapes learned by CNNs smoother and more stable, which 
brought about better generalization ability.
Motivated by the importance of feature map multiplication in AM and verifying
its effectiveness, we simply replace the feature map addition operator of ResNet with
feature map multiplication, and proposed FMMNet (Feature Map Multiplication Networks).
Formally, the output of a building block in ResNet was $\mathcal{F}(x) + x$ and that in
FMMNet was $\mathcal{F}(x) \times x$, where $+$ and $\times$ were the element-wise addition
and element-wise multiplication operator respectively. By comparing the performance of
ResNet and FMMNet, we verified the effectiveness of feature map multiplication.

\subsubsection{Implementation}
We modified wide ResNet structures into FMMNets by replacing the feature map addition operator
with feature map multiplication operator. 
They were evaluated and compared on four datasets, CIFAR-10, CIFAR-100,
STL-10, and TinyImageNet.
All the models were trained 
using momentum optimizer for 200 epochs with a momentum of 0.9.

\subsubsection{Result}

\begin{table}[ht]
	\centering
	\caption{The experimental results of ResNets and FMMNets.
		FMMNets outperformed ResNets on all datasets, without the increase of model parameters
		and computational costs. Especially on CIFAR-100 dataset, FMMNet outperformed ResNet
		by 0.57\%. It verified the effectiveness of feature map multiplication,
		and demonstrated that without a visual explanation,
		feature map multiplication could improve the performance of CNNs by learned smoother and more stable landscapes.
	}
	\renewcommand\arraystretch{1.2}
	\label{tab:fmmnet}
	\begin{tabular}{c|cccc}
		\hline 
		Datasets &CIFAR-10& CIFAR-100&STL-10 &TinyImageNet\\
		\hline
		ResNet &94.36
		&77.78
		& 81.39
		&58.57\\
		FMMNet&94.49&78.35&81.44 &58.84\\
		\hline	
	\end{tabular}
\end{table}

The results of ResNets and FMMNets were listed in table \ref{tab:fmmnet}.
FMMNets outperformed ResNets on all datasets, without the increase of model parameters
and computational costs. Especially on CIFAR-100 dataset, FMMNet outperformed ResNet
by 0.57\%. It verified the effectiveness of feature map multiplication,
and demonstrated that without a visual explanation,
feature map multiplication could improve the performance of CNNs by learned smoother and more stable landscapes.

However, more extensive experiments (not listed) showed that 
very deep FMMNets (more than 7 FMM building blocks)
suffered from the notorious problem
of exploding gradients, which hamper their convergence. We speculated that
the feature map multiplication operator affected the statistics of output feature maps,
and  the existing initialization and 
normalization strategies \cite{BN,He2015DelvingDI} designed for the feature map addition operator
were not designed for it.

\subsubsection{Conclusion}
As claimed in section \ref{sec:enlightenment},
AM benefited from the smooth and stable
landscapes brought by feature map multiplication, not the visual attention.
This section verified the effectiveness of feature map multiplication.
However, two questions also remained and could be explored by future works,
\begin{itemize}
	\item How the feature map multiplication operator affected the training process of CNNs?
	
	\item How to avoid the exploding gradients problem of FMMNets? 
\end{itemize}

\section{Conclusion}
This paper explored the effectiveness of attention mechanism. Firstly, according to
its widely accepted visual attention explanation, we compared the consistency
scores \cite{brandenburg2013comparing} between the attention weights
of features and their importance scores evaluated by saliency methods \cite{2019FullGradientRF,2017arXiv170603825S}.
It demonstrated that
there was only a weak consistency between them and
attention weights do not precisely express the importance of features.

Secondly, we verified that feature map multiplication in attention mechanism building blocks
bringing about high order non-linearity to CNNs, was the key for the effectiveness of attention
mechanism. We visualized and compared the landscapes learned by attention equipped CNNs
and vanilla CNNs and proposed that feature map multiplication made the learned landscapes smoother
and more stable near real samples compared to vanilla CNNs. Moreover, we conjectured and verified
the regularization role attention mechanism played on the learned landscapes of CNNs by using it
in conjunction with the augmentation method, mixup.

Finally, motivated by the effectiveness of feature map multiplication bringing the
high order non-linearity to CNNs. We designed FMMNet by simply replacing the 
feature map addition operator with feature map multiplication operator,
and verified its effectiveness on four datasets.



\ifCLASSOPTIONcaptionsoff
 \newpage
\fi

{\small
	\normalem

}

\begin{IEEEbiographynophoto}{Yong Li}
received MS in Applied Mathematics with Prof. Gerald Misiolek, and MSEE and PhD
in Electrical Engineering with Prof. Robert L. Stevenson, all from the University of Notre Dame.
He is now with School of Electronic Engineering, Beijing Univ. of Posts and Teles. His research is 
focused on machine vision, multispectral images, deep learning, and differential geometry.
\end{IEEEbiographynophoto}

\begin{IEEEbiographynophoto}{Xiang Ye}
	received the B.S. degree in the School of Engineering and Technology, China University of Geosciences (Beijing)
 in 2015. He is currently working towards the Ph.D degree in the Beijing Key Laboratory of Work Safety
Intelligent Monitoring, the Department of EE, BUPT. His research interests including computer vision,
object detection, and deep learning.
\end{IEEEbiographynophoto}

\end{document}